\documentclass[11pt]{article}

\usepackage[preprint]{acl}

\usepackage{times}
\usepackage{latexsym}

\usepackage[T1]{fontenc}

\usepackage[utf8]{inputenc}

\usepackage{microtype}

\usepackage{inconsolata}
\usepackage{listings}
\usepackage{graphicx}
\usepackage{amsmath}
\usepackage{dsfont}
\usepackage{amssymb}
\usepackage{booktabs}   %
\usepackage{multirow}   %
\usepackage{tabularx}
\usepackage{enumitem}
\setlist{leftmargin=1.2em, itemsep=1pt, parsep=0pt, topsep=3pt}
\usepackage{graphicx}   %
\usepackage[table]{xcolor}

\usepackage[most]{tcolorbox}

\definecolor{bestcell}{HTML}{D5F5E3}
\definecolor{secondcell}{HTML}{EBF5FB}
\definecolor{oursrow}{HTML}{FFF9E6}

\definecolor{keycolor}{HTML}{FFF3E0}
\definecolor{keyframe}{HTML}{E65100}
\definecolor{omitcolor}{HTML}{F5F5F5}
\definecolor{omitframe}{HTML}{BDBDBD}
\definecolor{resultcolor}{HTML}{E8F5E9}
\definecolor{resultframe}{HTML}{2E7D32}
\tcbset{
  promptstyle/.style={enhanced, breakable, colback=black!2, colframe=orange!60!black,
    boxrule=0.5pt, arc=1.0mm, colbacktitle=brown!80!black, coltitle=white,
    fonttitle=\small\bfseries, before skip=6pt, after skip=8pt},
  auditstyle/.style={colback=white, colframe=gray!65, boxrule=0.6pt, arc=1.5mm,
    coltitle=black, colbacktitle=gray!15, left=5pt, right=5pt, top=5pt, bottom=5pt},
  casestyle/.style={enhanced, colback=keycolor, colframe=keyframe, boxrule=0.6pt, arc=1.5mm,
    fonttitle=\small\bfseries, coltitle=white, colbacktitle=keyframe,
    left=4pt, right=4pt, top=2pt, bottom=2pt},
  resultstyle/.style={casestyle, colback=resultcolor, colframe=resultframe, colbacktitle=resultframe},
  omitstyle/.style={enhanced, colback=omitcolor, colframe=omitframe, boxrule=0.4pt, arc=1.0mm,
    left=4pt, right=4pt, top=1pt, bottom=1pt},
}
\newtcolorbox{promptbox}[2][]{promptstyle, title={#2}, #1}
\newtcolorbox{reviewquotetag}[1][]{promptstyle, #1}
\lstdefinestyle{promptlisting}{basicstyle=\ttfamily\scriptsize, breaklines=true,
  columns=fullflexible, keepspaces=true, breakindent=1.5em}

\newcommand{\methodname}{\texttt{ProReviewer}}

\title{ProReviewer: Evidence-Driven Investigation for Scientific Peer Review}

\author{Haishuo Fang\textsuperscript{1,2} \quad Yue Feng\textsuperscript{3} \quad Iryna Gurevych\textsuperscript{1,2}\\ 
  \textsuperscript{1}Ubiquitous Knowledge Processing Lab (UKP Lab), Technical University of Darmstadt \\
  \textsuperscript{2}National Research Center for Applied Cybersecurity ATHENE, Germany\\
  \textsuperscript{3}School of Computer Science, University of Birmingham \\
  \texttt{\href{https://www.ukp.tu-darmstadt.de}{www.ukp.tu-darmstadt.de}} \quad \texttt{\href{mailto:y.feng.6@bham.ac.uk}{y.feng.6@bham.ac.uk}}}

\begin{document}
\maketitle
\newcommand{\abstractfootnote}{\footnote{\raggedright Project page: \url{https://ukplab.github.io/arxiv2026-ProReviewer/}}}
\begin{abstract}
  Effective peer review requires adaptive investigation that responds to evidence emerging from each paper.
Existing LLM-based reviewers largely organize the review process around predefined workflows, leaving limited scope for emerging evidence to reshape investigation.
We propose \methodname{}, an evidence-driven review agent that formulates reviewing as a Markov Decision Process, with a structured \emph{review log} forming the core of its state.
At each step, the agent uses accumulated findings to decide what to inspect next and when to conclude.
As new evidence emerges, it updates the log by revising claim statuses, resolving open questions, and recording new findings.
Each point in the final review cites supporting log entries, making its judgments traceable to the recorded evidence.
On 1,000 held-out ICLR~2026 papers, the trained 8B \methodname{} achieves the highest average score among all compared methods, including those built on a 397B model; human evaluators prefer its reviews over all baselines on three of four review-quality dimensions.\ifdefined\abstractfootnote\abstractfootnote\fi

\end{abstract}

\section{Introduction}
The use of LLM-generated reviews in peer-review workflows remains debated~\citep{baumann2026stop,biswas2026aaai,neurips2026aireview}.
A key concern is whether these reviews result from careful investigation of each manuscript, rather than from a generic pass over its text.
\begin{figure}[htb!]
    \centering
    \includegraphics[width=1.0\columnwidth]{imgs/running\_exp.png}
    \caption{Illustrative example: \methodname{} checks an introduction claim (``robustness across domains'') against the experiments, finds only one domain evaluated, and records this evidence gap in its review log.}
    \label{fig:conceptual_overview}
\end{figure}
Studies show that LLM reviewers may accept authors' claims without scrutiny~\citep{du2024llms,ye2024peerreview}, overlook inconsistencies between claims and results~\citep{dycke2025automatic,li2025aspect}, and produce generic criticisms~\citep{li2025unveiling,qu2025claimcheck}.
These failures highlight the need to investigate preliminary impressions against evidence elsewhere in the manuscript.

However, existing approaches offer limited scope for such evidence-driven investigation.
Earlier systems follow predefined reviewing workflows~\cite{zhu2025deepreview,jin2024agentreview,idahl2025openreviewer}, which specify what to examine largely independently of the evidence found along the way.
Recent work makes investigation responsive to evidence by question decomposition~\cite{chang2025treereview} or verification agenda design~\cite{weng2025deepreview2}, but these changes remain within local parts of the fixed workflows.
This limits the ability to follow up on emerging concerns and reassess earlier judgments in response to new evidence, as human expert reviewers do.
Human expert reviewers connect evidence across sections, revisit earlier claims when inconsistencies surface, and decide what to inspect next based on what they have already found~\cite{willis2024peer}.
For example, when a claim in the introduction is contradicted by results in the experiments, a human reviewer would cross-check and flag the discrepancy (Figure~\ref{fig:conceptual_overview}), whereas a system with a fixed investigation path may accept the claim at face value and never revisit it.

This motivates us to explore whether peer review can be organized as an evidence-driven investigation process, in which accumulated findings continually guide what to investigate next.
In this paper, we introduce \methodname{}, a review agent whose investigation is driven by the evidence it accumulates.
To realize this, \methodname{} formulates peer review as a Markov Decision Process (MDP) whose state is represented by a structured \emph{review log}. A key distinction from existing agent paradigms is that the state in \methodname{} is not a record of interaction trajectory but a structured representation of the \textit{evolving evidentiary state of reviewing}, such as verified claims, unresolved questions. 
The agent uses this state to decide what to investigate next and when to conclude.
As new evidence emerges, the agent updates the log by revising claim statuses, resolving open questions, and recording new findings. This log thus serves a dual role: as the decision state, it guides the investigation according to current evidentiary needs; as an evidence trail, it preserves the sources and revision history of its entries. \methodname{} outputs both the final review and the resulting log, with each review point linked to its supporting entries, making the review judgments traceable to concrete evidence.

To instantiate our approach as a specialized review agent, we train \methodname{} through supervised fine-tuning, followed by GRPO~\citep{deepseek2025r1}.
To reduce data contamination risk, we construct a version-matched corpus of 5K ICLR~2025/2026 paper--review pairs, holding out 1K ICLR~2026 papers for testing (\S\ref{sec:setup}).
We evaluate \methodname{}'s review quality and audit the evidence underlying its judgments, complemented by a controlled evaluation of injected logical error detection.
On review quality, the trained 8B \methodname{} achieves the highest average score among all compared baselines, including those built on a 397B model (\S\ref{sec:main_results}). 
Human evaluators favor its reviews over those of all baselines on three of four review-quality dimensions (\S\ref{sec:human_eval}).
When applied as a prompting framework to the 397B model, \methodname{} further improves the average score to 0.60, compared with 0.55 for the strongest baseline on the same backbone (\S\ref{sec:prompting_strategy}).
Auditing the evidence trail, we find that 98.3\% of review points are supported by their cited log entries, and 87.2\% of the cited entries are faithful to the source manuscripts (\S\ref{sec:evidence_support}).
On a benchmark of injected logical errors~\citep{dycke2025automatic}, \methodname{} identifies 27\% of the errors in its final reviews, compared with 21\% for the strongest baseline (\S\ref{sec:counterfactual}).

Our contributions can be summarized as:
\newcounter{contribctr}
\begin{list}{\arabic{contribctr}.}{\usecounter{contribctr}\leftmargin=1em \itemindent=0pt \topsep=2pt \itemsep=1pt \parsep=0pt \partopsep=0pt}
    \item \methodname{}, an evidence-driven review agent formulated as an MDP, whose \emph{review log} both drives investigation and grounds the review.
    \item A curated, version-matched corpus of 5K ICLR~2025/2026 paper--review pairs for temporally separated training and evaluation.
    \item Experiments showing that \methodname{} improves review quality and injected-error detection, complemented by a human audit assessing the faithfulness and support of its evidence trails.
\end{list}

\section{Related Work}\label{sec:related}
\paragraph{LLM-based Review Generation.}
LLM-based reviewing spans direct prompting~\citep{robertson2023gpt4,liu2023reviewergpt}, specialized training~\citep{weng2025cycleresearcher,zeng2025reviewrl}, staged pipelines~\citep{zhu2025deepreview}, and multi-agent collaboration~\citep{jin2024agentreview,goyal2026scholarpeer,yamada2025ai}.
These systems largely define their workflows before any evidence is gathered.
Recent work further explores how evidence can be used in the review process.
TreeReview~\citep{chang2025treereview} decomposes reviewing into a question tree, where evidence only triggers local follow-up questions.
Without a shared evidence state, findings are not reused across branches, and cost grows with tree size and depth.
Concurrent work, DeepReviewer~2.0~\citep{weng2025deepreview2}, builds an evidence ledger and a verification agenda, with evidence guiding which passages to read during the verification stage while the initially constructed agenda remains fixed.
\methodname{} instead maintains a shared, evolving review log across the investigation, using it to guide subsequent actions and revise earlier judgments, and to ground the final review.

\paragraph{Agentic Reasoning and Memory.}
Autonomous LLM agents have demonstrated strong performance in diverse tasks such as web navigation~\citep{nakano2021webgpt,zhou2024webarena}, software engineering~\citep{jimenez2024swebench,wang2025openhands}, and neuro-symbolic reasoning~\citep{fang-etal-2024-dara,huang-etal-2025-targa}.
These agents follow the ReAct paradigm~\citep{yao2023react}, where state is implicitly represented by an append-only interaction trajectory interleaved with free-form reasoning.
While effective for many tasks, this representation is not designed to support a structured evidence state with explicit status updates.
Peer review instead requires maintaining explicit objects (e.g., claims and questions) whose status evolves as new evidence is collected.
To bridge this gap, \methodname{} proposes a structured review log as the agent state, enabling explicit tracking and updating of evidence throughout the review process.
This review log serves not only as a repository of intermediate information but also as the decision state in the MDP formulation, which complements recent work on agent memory~\citep{shinn2023reflexion,park2023generative,hu2025memory,yang2025memoryr1} for peer review.

\section{Method}
\begin{figure*}
    \centering
    \includegraphics[width=0.94\textwidth]{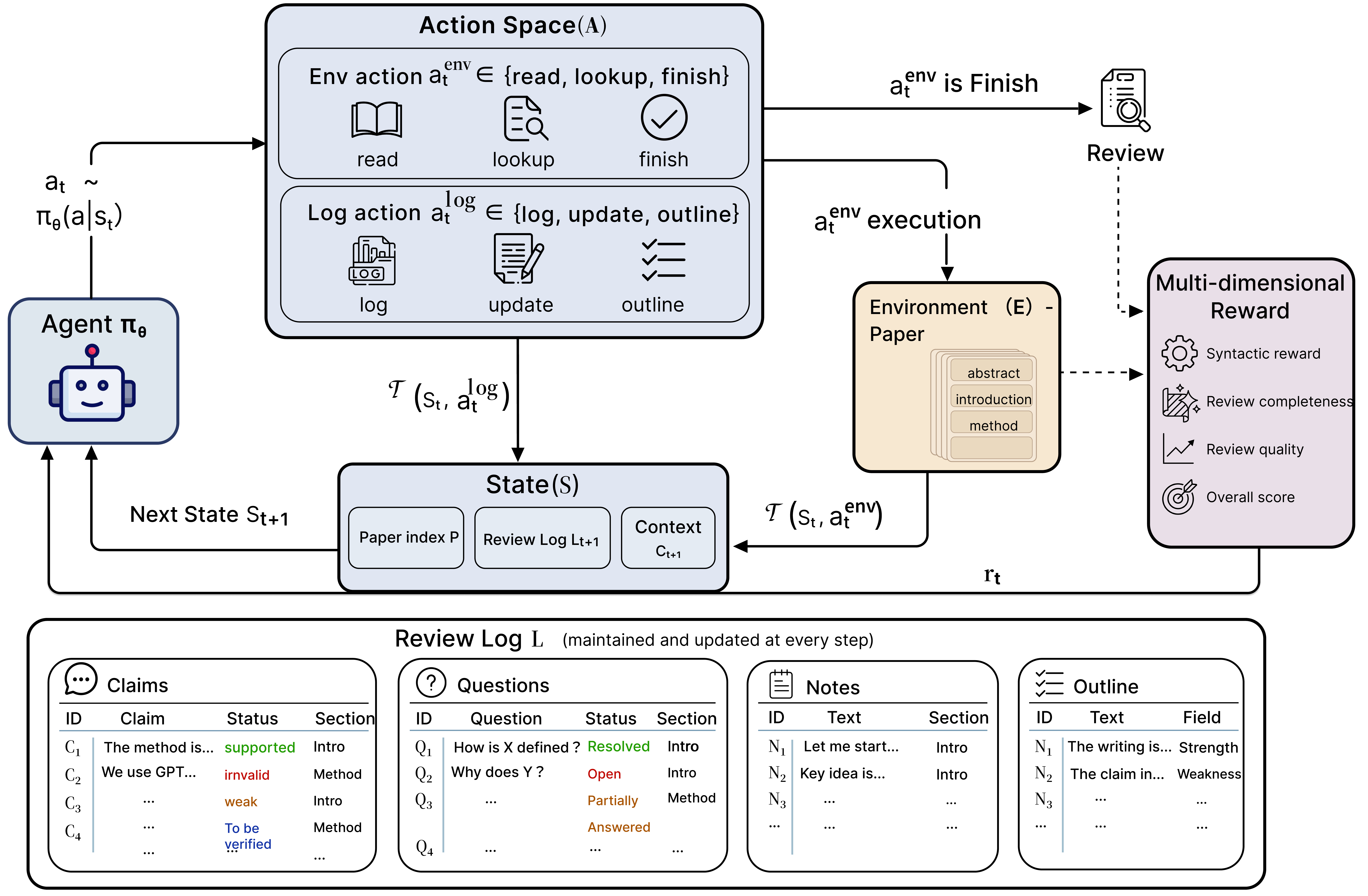}
    \caption{Interaction loop of \methodname{}. At time step $t$, the agent $\pi_\theta$ observes state $s_t$ (paper index, review log, and context) and samples an action $a_t=(a_t^{\mathrm{env}}, a_t^{\mathrm{log}})$. $a_t^{\mathrm{env}}$ retrieves paper content, and $a_t^{\mathrm{log}}$ updates the review log. A multi-component reward produces $r_t$, and the system transitions to $s_{t+1}$ until termination.}
    \label{fig:reviewer}
\end{figure*}

In this section, we show how \methodname{} uses accumulated evidence to guide manuscript investigation and turns its findings into a traceable review (Figure~\ref{fig:reviewer}).
We first describe how the agent examines the manuscript and updates its evidence state within an MDP (\S\ref{sec:mdp}).
We then explain how the review log organizes this evidence and links it to review points (\S\ref{sec:review_log}), before presenting the reward design and training procedure used to instantiate the framework as a trained agent (\S\ref{sec:instantiation}).
A concrete case study is provided in Appendix~\ref{app:case_study}.

\subsection{Reviewing as a Markov Decision Process}
\label{sec:mdp}
Reviewing in \methodname{} unfolds as a sequence of decisions: at each step, based on its current context and review log, the agent chooses what to examine next and what to record or revise in the log.
We formalize this as $\mathcal{M} = (\mathcal{S}, \mathcal{A}, \mathcal{T}, \mathcal{E}, \mathcal{R})$.

\paragraph{State $\mathcal{S}$.}
The state specifies the information available to the agent at each decision step.
Each state $s_t = (\mathcal{C}_t, \mathcal{L}_t, \mathcal{P})$ consists of
(1)~the \textbf{current context} $\mathcal{C}_t$, the most recent content from the paper (e.g., the text of a section);
(2)~the \textbf{review log} $\mathcal{L}_t$, the evidence entries accumulated so far and the developing review outline (\S\ref{sec:review_log}); and
(3)~the \textbf{paper index} $\mathcal{P}$, the title and table of contents, which the agent uses to navigate the paper.
Since $\mathcal{C}_t$ retains only the latest observation, the review log is what carries accumulated evidence across steps and guides each subsequent decision.

\paragraph{Action $\mathcal{A}$.}
At each step, the agent both gathers information from the manuscript and records what it has learned.
Accordingly, each action $a_t=(a_t^{\mathrm{env}},a_t^{\mathrm{log}})$ pairs one environment action with a list of log operations, both conditioned on $s_t$:
the log operations record evidence from the latest observation in $\mathcal{C}_t$, while the environment action determines what to examine next.
\textbf{Environment actions} include read a section (\textit{read\_section}), look up a specific element such as a table, or equation (\textit{look\_up}), or end the review (\textit{finish}).
\textbf{Log operations} add new entries (\textit{log}), revise existing ones (\textit{update}), and draft the review outline (\textit{outline}); see \S\ref{sec:review_log}.
Because the policy selects both types of actions at every step, the agent can interleave reading, recording, and writing as the evidence demands, rather than following fixed stages.
In this work, we restrict the action space to the manuscript itself, excluding external retrieval, so as to evaluate our core design in isolation.
The formulation is modular: actions such as literature search for novelty assessment can be added without changing the core architecture, as demonstrated by the prototype in Appendix~\ref{app:novelty_search}.
Full action schemas are given in Appendix~\ref{app:action_schema} (Table~\ref{tab:action_schema}).

\paragraph{Transition $\mathcal{T}$ and Environment $\mathcal{E}$.}
The environment $\mathcal{E}$ is the manuscript under review, so the transition $s_{t+1}=\mathcal{T}(s_t,a_t)$ is deterministic.
At each step, the log operations record what the agent has learned from $\mathcal{C}_t$ into $\mathcal{L}_{t+1}$ (invalid operations are rejected; \S\ref{sec:review_log}), and the environment action either fetches new content as $\mathcal{C}_{t+1}$ or ends the episode.
The paper index $\mathcal{P}$ stays fixed.

\paragraph{Reward $\mathcal{R}$.}
We define a multi-dimensional reward to cover both action validity at each step and the overall quality of the final review, which is described in \S\ref{sec:reward}.

\paragraph{Trajectory.}
A review episode is a trajectory $\tau = (s_0, a_0, s_1, \ldots, s_T)$ generated by the policy $\pi_\theta$.
The agent starts with an empty context and review log, $s_0 = (\varnothing, \varnothing, \mathcal{P})$, and at each step samples $a_t \sim \pi_\theta(\cdot \mid s_t)$ to obtain $s_{t+1} = \mathcal{T}(s_t, a_t)$.
The episode ends when the agent issues \textit{finish} or reaches the step limit $T_{\max}$, after which the outline entries in $\mathcal{L}_T$ are rendered into the final review.
Rewards used for training are described in \S\ref{sec:reward}.

\subsection{State Design: Review Log}
\label{sec:review_log}
The \textbf{review log} $\mathcal{L}$ connects investigation with review writing through two components: evidence entries that track claims, questions, and notes, and a review outline where stores the final review whose strengths and weaknesses cite evidence entries.
By reading the log, the agent sees which claims still need checking and which questions remain open, and builds review points from the evidence collected so far.

\paragraph{Evidence entries.}
The log maintains three types of evidence entries, each with a unique identifier and a provenance record of its source section, creation step, and update history:
\textbf{claims} $\{C_1, C_2, \ldots\}$, assertions from the paper with a verification status (\texttt{to\_be\_verified}, \texttt{supported}, \texttt{weak}, or \texttt{invalid});
\textbf{questions} $\{Q_1, Q_2, \ldots\}$, open issues the agent raises during reading, with a resolution status (\texttt{open}, \texttt{partially\_answered}, or \texttt{resolved});
and \textbf{notes} $\{N_1, N_2, \ldots\}$, intermediate findings and observations.
The agent adds entries via \textit{log} and updates their status via \textit{update} as new evidence emerges (e.g., marking a claim as \texttt{supported} after finding corroboration later).

\paragraph{Review outline.}
The outline stores the developing summary, strengths, weaknesses, questions for the authors, and overall score.
Through \textit{outline}, the agent adds review points as evidence accumulates, and each strength and weakness must cite the entries it rests on; points without evidence tags or with references to non-existent entries are rejected.
For example, the weakness \texttt{"Limited baseline comparison [C1, Q2, N5]"} links the critique to claim C1, question Q2, and note N5.
This mechanism makes the final review traceable: \methodname{} outputs the review log alongside the review, linking every strength and weakness to its cited log entries.

\subsection{Instantiating the Review Agent}
\label{sec:instantiation}
The MDP formulation makes the investigation policy learnable, so we instantiate the framework as a specialized review agent through supervised fine-tuning and reinforcement learning.

\subsubsection{Reward Design}
\label{sec:reward}

The reward combines four components: step-level action validity and three trajectory-level scores for completeness, content quality, and score alignment.

\paragraph{Step-level Validity.}
To teach the agent correct action invocation, the validity reward $r^{\text{val}}_t$ provides immediate feedback on action validity at each step:
  \begin{equation}
  r^{\text{val}}_t = -\mathds{1}\bigl[\lambda_{\text{form}} \lor \lambda_{\text{exec}}
  \lor \lambda_{\text{ground}} \lor \lambda_{\text{cov}}\bigr] \in \{-1, 0\}
  \end{equation}
where \textit{formatting errors} ($\lambda_{\text{form}}$) indicate schema violations (e.g., malformed JSON),
\textit{execution errors} ($\lambda_{\text{exec}}$) indicate invalid action names or arguments (e.g., querying a non-existent section),
\textit{evidence-reference errors} ($\lambda_{\text{ground}}$) penalize references to evidence not present in the agent's log, and
\textit{coverage errors} ($\lambda_{\text{cov}}$) penalize a \textit{finish} action issued before all main sections of the paper have been read.

\paragraph{Review Completeness.}
We define review completeness as $r^{\text{comp}} = \frac{1}{4} \sum_{i=1}^{4} \mathds{1}[\text{check}_i \text{ satisfied}]$.
The four checks verify: (1)~a summary is present, (2)~at least one strength, (3)~at least one weakness, and (4)~an overall score.

\paragraph{Review Content Quality.}
Beyond structural completeness, a high-quality review must demonstrate substantive engagement with the paper's technical content~\citep{zhu2025deepreview,garg-etal-2025-revieweval,goyal2026scholarpeer}.
We capture this with two rubric-based LLM-as-a-judge scores: \textit{technical depth} ($r^{\text{depth}}$), rewarding engagement with methodological and experimental detail, and \textit{grounding} ($r^{\text{grd}}$), rewarding critiques anchored in concrete paper content.
They are combined as $r^{\text{qual}} = \alpha \cdot r^{\text{depth}} + (1-\alpha) \cdot r^{\text{grd}}$, with $\alpha = 0.5$.

\paragraph{Score Alignment.}
Beyond textual feedback, peer reviews typically include a quantitative assessment.
We encourage the agent to align its score $\hat{s}$ with the human reviewer average $\bar{s}$:
\begin{equation}
r^{\text{scr}} = \max\bigl(0,\, 1 - |\hat{s} - \bar{s}| / \kappa\bigr)
\label{eq:score_align}
\end{equation}
where $\kappa$ is the rating scale range (e.g., $\kappa = 9$ for scores on a 1--10 scale).

\paragraph{Total Reward.}
Optimizing multiple quality objectives has been studied in text generation~\citep{pasunuru-bansal-2018-multi,ryu-etal-2024-multi}.
We combine the components as a weighted sum: the reward assigned to step $t$ is $r_t = w_{\text{val}} \cdot r^{\text{val}}_t + \sum_{k \in K} w_k \cdot r^k$, where $K=\{\text{comp},\text{qual},\text{scr}\}$.
Step-level validity is assessed when actions are executed; trajectory-level rewards are computed at episode termination and broadcast uniformly to all steps.
This simple weighted sum yields strong performance in our setting; the specific weights and comparison with equal weighting are given in Appendix~\ref{app:hyperparams} (Table~\ref{tab:reward_weighting}).

\subsubsection{Training}\label{sec:train}
We adopt a two-stage training pipeline to obtain an agent that produces valid actions and explores efficiently within a limited interaction budget.
First, we perform supervised fine-tuning (SFT) on interaction trajectories distilled from a strong teacher model (Qwen3.5-397B-A17B; see Appendix~\ref{app:data} for dataset construction).
Second, starting from the SFT checkpoint, we apply GRPO reinforcement learning with a two-phase curriculum: Phase~1 trains with only deterministic, rule-based rewards (step-level validity, review completeness, and score alignment).
For Phase~2, we additionally include LLM-judge-based content quality rewards while retaining all Phase~1 rewards.

\section{Experiments}
\subsection{Experiment Setup}\label{sec:setup}
\paragraph{Dataset.}
To facilitate reproducible experiments, following prior work~\citep{zhu2025deepreview,weng2025cycleresearcher,goyal2026scholarpeer}, we use peer-review data from the International Conference on Learning Representations (ICLR), which is publicly available and covers a wide range of AI research topics in this fast-paced field.
We collected submissions across two ICLR conference cycles (2025--2026),
carefully matching each paper's initial submission with its corresponding reviews and initial scores
to ensure version alignment.
After filtering for review completeness,
the final corpus comprises 5{,}011 papers:
4{,}011 ICLR 2025 papers for training and validation,
and 1{,}000 ICLR 2026 papers for evaluation.
The temporal separation ensures that test papers were published after base model training,
mitigating potential data contamination
(see Appendix~\ref{app:data} for full details).
\paragraph{Baselines.}
We compare against three categories of baselines that span different paradigms in automated review generation.
(1) For~\textbf{prompt-based methods}, we include three representative methods: AgentReview~\citep{jin2024agentreview}, AI-Scientist-v2~\citep{yamada2025ai}, and TreeReview~\citep{chang2025treereview}.
We evaluate each method with Llama3.1-8B, Qwen3-8B, and Qwen3.5-397B-A17B, as well as the commercial model Gemini-3.1-flash-lite (Table~\ref{tab:main_results}).
(2)~\textbf{Supervised fine-tuning.} We include CycleReviewer~\citep{weng2025cycleresearcher} and DeepReview~\citep{zhu2025deepreview}, which represent the state of the art in SFT-based review generation.
(3)~\textbf{Reinforcement learning.} We reimplement \textit{ReviewRL}~\citep{zeng2025reviewrl}, denoted \textit{ReviewRL~(Reimpl.)} as the original implementation is not publicly available.
\paragraph{Implementation Details.}
Fine-tuned methods are built on Qwen3-8B~\citep{yang2025qwen3} and Llama3.1-8B~\citep{grattafiori2024llama} to test generalization across model families, with the backbones of each method listed in Table~\ref{tab:main_results}.
To ensure a fair comparison, all fine-tuned baselines are trained on the same training data and backbones as \methodname{}.
For RL training, we use GPT-OSS-120B~\citep{agarwal2025gpt} as the LLM judge for reward computation.
Implementation details are provided in Appendix~\ref{app:hyperparams}.

\paragraph{Inference Cost.}
\methodname{} makes multiple LLM calls per paper and uses more tokens than single-pass or multi-stage methods, but far fewer than TreeReview, and it retains its advantage under an equal token budget.
The full comparison of token usage, inference time, monetary cost, and the equal-budget results is reported in Appendix~\ref{app:complexity}.

\paragraph{Evaluation Protocol.}
We first evaluate the quality of final reviews (\S\ref{sec:main_results}), then audit whether the underlying evidence faithfully reflects the manuscript and supports those judgments (\S\ref{sec:evidence_support}).
Finally, we test whether the reviews detect injected logical errors with known ground truth (\S\ref{sec:counterfactual}).

\subsection{Review Quality}\label{sec:main_results}
\begin{table*}[htb!]
  \centering
  \definecolor{bestcell}{HTML}{D5F5E3}
  \definecolor{secondcell}{HTML}{EBF5FB}
  \providecommand{\best}[1]{\cellcolor{bestcell}\textbf{#1}}
  \providecommand{\second}[1]{\cellcolor{secondcell}#1}
  \providecommand{\ms}[2]{#1{\scriptsize\textcolor{black!90}{$\pm$#2}}}
  \setlength{\tabcolsep}{4pt}
  \newlength{\maintabdashw}
  \newcommand{\maintabular}{%
  \begin{tabular}{@{}llcccccccccccc@{}}
  \toprule
  \multirow{2}{*}{\textbf{Method}} & \multirow{2}{*}{\textbf{Model}} &
  \multicolumn{2}{c}{\textbf{Actionability}} &
  \multicolumn{2}{c}{\textbf{Grounding}} &
  \multicolumn{2}{c}{\textbf{Technical Depth}} &
  \multicolumn{2}{c}{\textbf{Verifiability}} &
  \multicolumn{2}{c}{\textbf{Score Alignment}} &
  \multicolumn{2}{c}{\textbf{Avg.}} \\
  \cmidrule(lr){3-4} \cmidrule(lr){5-6} \cmidrule(lr){7-8} \cmidrule(lr){9-10} \cmidrule(lr){11-12} \cmidrule(lr){13-14}
  & & avg@4 & best@4 & avg@4 & best@4 & avg@4 & best@4 & avg@4 & best@4 & avg@4 & best@4 & avg@4 & best@4 \\
  \midrule

  \multicolumn{14}{c}{\textit{Prompt-based methods}} \\
  \midrule

  AgentReview & Llama3.1-8B & \ms{.13}{.08} & \ms{.14}{.09} & \ms{.12}{.09} & \ms{.13}{.10} & \ms{.13}{.10} & \ms{.14}{.10} & \ms{.06}{.05} & \ms{.07}{.06} & \ms{.64}{.14} & \ms{.66}{.15} & \ms{.21}{.09} & \ms{.23}{.10} \\
   & Qwen3-8B & \ms{.30}{.08} & \ms{.35}{.08} & \ms{.32}{.12} & \ms{.40}{.11} & \ms{.21}{.09} & \ms{.26}{.09} & \ms{.13}{.09} & \ms{.18}{.10} & \ms{.63}{.13} & \ms{.66}{.13} & \ms{.32}{.10} & \ms{.37}{.10} \\
   & Gemini-3.1-FL & \ms{.29}{.11} & \ms{.37}{.12} & \ms{.36}{.13} & \ms{.46}{.13} & \ms{.44}{.12} & \ms{.50}{.11} & \ms{.25}{.12} & \ms{.33}{.12} & \ms{.70}{.14} & \ms{.77}{.13} & \ms{.41}{.13} & \ms{.49}{.12} \\
   & Qwen3.5-397B & \ms{.28}{.11} & \ms{.35}{.12} & \ms{.46}{.15} & \ms{.57}{.14} & \second{\ms{.45}{.13}} & \ms{.52}{.12} & \ms{.23}{.10} & \ms{.29}{.10} & \ms{.78}{.13} & \ms{.79}{.13} & \ms{.44}{.12} & \ms{.50}{.12} \\

  \cmidrule{1-14}

  AI-Scientist-v2 & Llama3.1-8B & \ms{.17}{.11} & \ms{.25}{.11} & \ms{.16}{.13} & \ms{.26}{.16} & \ms{.14}{.10} & \ms{.20}{.10} & \ms{.07}{.05} & \ms{.09}{.06} & \ms{.74}{.14} & \ms{.81}{.13} & \ms{.25}{.11} & \ms{.32}{.11} \\
   & Qwen3-8B & \ms{.21}{.07} & \ms{.26}{.08} & \ms{.21}{.10} & \ms{.28}{.11} & \ms{.21}{.09} & \ms{.27}{.09} & \ms{.09}{.05} & \ms{.12}{.06} & \ms{.68}{.14} & \ms{.69}{.14} & \ms{.28}{.09} & \ms{.32}{.09} \\
   & Gemini-3.1-FL & \ms{.26}{.11} & \ms{.34}{.12} & \ms{.31}{.13} & \ms{.41}{.13} & \ms{.38}{.14} & \ms{.47}{.13} & \ms{.19}{.11} & \ms{.26}{.12} & \ms{.73}{.13} & \ms{.81}{.13} & \ms{.37}{.13} & \ms{.46}{.13} \\
   & Qwen3.5-397B & \ms{.28}{.11} & \ms{.36}{.12} & \ms{.44}{.16} & \ms{.55}{.15} & \second{\ms{.45}{.14}} & \ms{.52}{.12} & \ms{.25}{.09} & \ms{.31}{.09} & \ms{.83}{.12} & \ms{.83}{.12} & \ms{.45}{.13} & \ms{.51}{.12} \\

  \cmidrule{1-14}

  TreeReview & Llama3.1-8B & \ms{.22}{.13} & \ms{.33}{.12} & \ms{.27}{.16} & \ms{.39}{.14} & \ms{.19}{.12} & \ms{.25}{.13} & \ms{.07}{.11} & \ms{.16}{.12} & \ms{.70}{.14} & \ms{.74}{.15} & \ms{.29}{.13} & \ms{.37}{.13} \\
   & Qwen3-8B & \ms{.16}{.11} & \ms{.28}{.10} & \ms{.39}{.14} & \ms{.54}{.12} & \ms{.28}{.13} & \ms{.41}{.08} & \ms{.17}{.12} & \ms{.28}{.07} & \ms{.65}{.20} & \ms{.78}{.08} & \ms{.33}{.14} & \ms{.46}{.09} \\
   & Gemini-3.1-FL & \ms{.29}{.12} & \ms{.39}{.13} & \ms{.55}{.13} & \ms{.60}{.13} & \ms{.40}{.12} & \ms{.48}{.13} & \second{\ms{.44}{.10}} & \ms{.48}{.10} & \ms{.78}{.13} & \ms{.83}{.12} & \ms{.49}{.12} & \ms{.56}{.12} \\
   & Qwen3.5-397B & \ms{.35}{.36} & \ms{.40}{.33} & \best{\ms{.66}{.24}} & \second{\ms{.72}{.22}} & \ms{.42}{.22} & \ms{.50}{.21} & \best{\ms{.51}{.15}} & \best{\ms{.53}{.14}} & \ms{.80}{.18} & \ms{.84}{.17} & \second{\ms{.55}{.23}} & \ms{.60}{.21} \\

  \midrule
  \multicolumn{14}{c}{\textit{Fine-tuned models}} \\
  \midrule

  CycleReviewer & Llama3.1-8B & \second{\ms{.39}{.21}} & \second{\ms{.49}{.19}} & \ms{.25}{.18} & \ms{.34}{.17} & \ms{.26}{.15} & \ms{.33}{.16} & \ms{.15}{.18} & \ms{.23}{.22} & \ms{.84}{.13} & \ms{.87}{.13} & \ms{.38}{.17} & \ms{.45}{.17} \\
   & Qwen3-8B & \ms{.31}{.19} & \ms{.37}{.19} & \ms{.28}{.19} & \ms{.36}{.20} & \ms{.21}{.14} & \ms{.27}{.15} & \ms{.17}{.18} & \ms{.21}{.20} & \ms{.84}{.14} & \ms{.84}{.20} & \ms{.36}{.17} & \ms{.41}{.19} \\
  DeepReview & Llama3.1-8B & \ms{.34}{.19} & \ms{.47}{.19} & \ms{.35}{.21} & \ms{.50}{.20} & \ms{.26}{.15} & \ms{.36}{.17} & \ms{.24}{.20} & \ms{.33}{.21} & \ms{.85}{.12} & \best{\ms{.92}{.10}} & \ms{.41}{.18} & \ms{.51}{.17} \\
   & Qwen3-8B & \ms{.35}{.20} & \ms{.46}{.17} & \ms{.22}{.18} & \ms{.33}{.18} & \ms{.23}{.15} & \ms{.32}{.16} & \ms{.13}{.17} & \ms{.23}{.21} & \ms{.85}{.11} & \second{\ms{.90}{.10}} & \ms{.35}{.16} & \ms{.45}{.17} \\
  ReviewRL (Reimpl.) & Qwen3-8B & \ms{.33}{.17} & \second{\ms{.49}{.14}} & \ms{.51}{.15} & \ms{.61}{.13} & \ms{.36}{.16} & \ms{.47}{.11} & \ms{.38}{.13} & \ms{.43}{.12} & \best{\ms{.89}{.12}} & \ms{.89}{.08} & \ms{.49}{.14} & \ms{.58}{.12} \\
  \noalign{\vskip 1.5pt\hbox to \maintabdashw{\color{black!55}\leaders\hbox to 3pt{\hss\rule[0pt]{1.8pt}{0.4pt}\hss}\hfill}\vskip 2pt}
  \textbf{\methodname\ (ours)} & Llama3.1-8B & \ms{.30}{.13} & \ms{.40}{.13} & \ms{.56}{.19} & \ms{.70}{.17} & \ms{.44}{.15} & \second{\ms{.54}{.13}} & \ms{.40}{.16} & \second{\ms{.50}{.14}} & \best{\ms{.89}{.08}} & \second{\ms{.90}{.08}} & \ms{.52}{.14} & \second{\ms{.61}{.13}} \\
   & Qwen3-8B & \best{\ms{.46}{.13}} & \best{\ms{.55}{.13}} & \second{\ms{.64}{.16}} & \best{\ms{.76}{.12}} & \best{\ms{.48}{.13}} & \best{\ms{.57}{.12}} & \ms{.40}{.13} & \ms{.48}{.12} & \second{\ms{.88}{.11}} & \second{\ms{.90}{.11}} & \best{\ms{.57}{.13}} & \best{\ms{.65}{.12}} \\

  \bottomrule
  \end{tabular}
  }
  \setlength{\maintabdashw}{0pt}\sbox0{\maintabular}\setlength{\maintabdashw}{\wd0}
  \resizebox{\textwidth}{!}{\maintabular}
  \caption{Averaged over three LLM judges, as mean{\scriptsize$\pm$std} on $[0,1]$ over four reviews per paper; RevUtil does not support Technical Depth, so that column is averaged over the other two judges (per-judge results in Appendix~\ref{app:per_judge}). Qwen3.5-397B = Qwen3.5-397B-A17B; FL = flash-lite. \colorbox{bestcell}{\textbf{Green}}: best; \colorbox{secondcell}{Blue}: second-best.}
  \label{tab:main_results}
\end{table*}
A useful review should help authors understand what to improve (\textit{Actionability}), where the issue arises (\textit{Grounding}), how deeply it engages with the technical substance (\textit{Technical Depth}), and why the critique is justified (\textit{Verifiability})~\cite{sadallah2025good,garg-etal-2025-revieweval,zhu2025deepreview}.
Following the review utility framework of \citet{sadallah2025good}, we score each dimension on a 1--5 rubric and normalize scores to [0,1].
Besides, we also report \textit{Score Alignment}, which measures the calibration of the overall rating (Eq.~\ref{eq:score_align}).

We evaluate four review-quality dimensions using both LLM-as-a-judge and human evaluation.
To reduce judge bias~\citep{zheng2023judging}, we aggregate scores from three diverse evaluators that are not used as base models in any baseline: GPT-5.4 nano, DeepSeek-V4 flash~\citep{deepseekai2026deepseekv4}, and RevUtil~\citep{sadallah2025good}, a review-quality evaluator fine-tuned on human annotations.
For each paper, we generate four independent reviews and report both average (avg@4) and best-of-four (best@4) scores.

\subsubsection{Automatic Evaluation Results}\label{sec:auto_eval}
As shown in Table~\ref{tab:main_results}, \methodname{} (Qwen3-8B) achieves the highest overall scores in both avg@4 (0.57) and best@4 (0.65), and remains first under within-judge score standardization (Appendix~\ref{app:per_judge}).
Under avg@4, it outperforms all prompt-based baselines on \textit{Actionability}, \textit{Technical Depth}, and \textit{Score Alignment}, including those built on the 397B backbone.
It performs comparably to TreeReview (397B) on \textit{Grounding} (0.64 vs.\ 0.66) but trails it on \textit{Verifiability} (0.40 vs.\ 0.51), which we analyze below.
Among fine-tuned methods, \methodname{} leads on all four rubric dimensions, with the largest margins on \textit{Grounding} (0.64 vs.\ 0.51) and \textit{Technical Depth} (0.48 vs.\ 0.36).
Notably, \methodname{} outperforms ReviewRL~(Reimpl.) (0.57 vs.\ 0.49), despite both methods sharing the same backbone, training stages, and reward functions, suggesting that the gains arise primarily from our architecture design rather than the training recipe alone.

The largest shortfall relative to TreeReview (397B) is in \textit{Verifiability}.
However, this gap reverses in the same-backbone prompting comparison: \methodname{} reaches 0.59, surpassing TreeReview on Qwen3.5-397B-A17B (0.51; \S\ref{sec:prompting_strategy}).
This suggests that the gap mainly stems from backbone capacity rather than the framework.

\subsubsection{Human Evaluation Results}\label{sec:human_eval}

\begin{figure}[!h]
\centering
\includegraphics[width=\columnwidth]{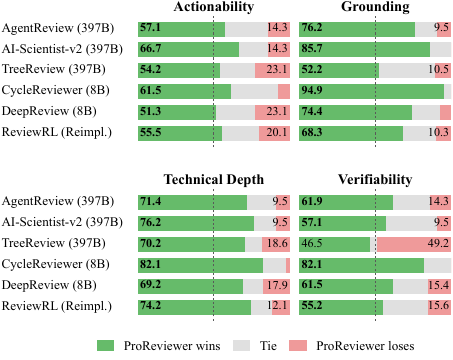}
\caption{Human pairwise evaluation against each baseline. 397B = Qwen3.5-397B-A17B; 8B = Qwen3-8B.}
\label{fig:human_eval}
\end{figure}

To validate the automatic evaluation, five reviewers experienced in reviewing for AI conferences perform blind pairwise comparisons between \methodname{} and each baseline on 50 randomly sampled test papers, across all quality dimensions (protocol in Appendix~\ref{app:human_eval}).
As shown in Figure~\ref{fig:human_eval}, the results mirror the automatic evaluation: \methodname{} is preferred in every matchup on Actionability, Grounding, and Technical Depth (win rates of 51\%--95\%), and on Verifiability against all baselines except TreeReview (46.5\% vs.\ 49.2\%).
The margins are generally largest on \emph{Grounding} and \emph{Technical Depth}, the dimensions most directly shaped by evidence-driven investigation.
The full Bradley--Terry ranking is given in Appendix~\ref{app:human_eval_bt}.

\subsection{Evidence Trail Audit}\label{sec:evidence_support}
One benefit of \methodname{} is that the evidence gathered during investigation is explicitly recorded in the review log and linked to the final reviews.
Although traceability alone does not establish correctness,
it enables human experts to audit the evidence underlying these judgments, supporting more trustworthy LLM reviewing.
We therefore audit whether the cited log entries support the corresponding review points (\textit{review-to-log support}), and whether those entries faithfully reflect the manuscript (\textit{log-to-paper faithfulness}).
Two human reviewers assess 120 review points from 30 randomly sampled papers, covering 172 cited log entries; the protocol and inter-annotator agreement are reported in Appendix~\ref{app:evidence_audit}.

Overall, 84.2\% of review points satisfy both criteria, 98.3\% are supported by their cited log entries, and 87.2\% of cited entries are faithful to the manuscript. 
These results suggest that more failures happen when extracting or interpreting manuscript evidence rather than in grounding review judgments in the log.
We identify two main failure modes: table misreadings and unsupported interpretations of reported results. 
Table misreadings arise when the agent reads a number from the wrong row or column of a table; 60\% of them occur in complex tables with multi-level or merged column headers.
Unsupported interpretations arise when the agent correctly records numerical results but adds an explanation the manuscript never states.
We also examine \textit{update} operations under the same log-to-paper criterion: updates are frequent (86.9\% of cited claims and questions) and largely reliable, with errors concentrated in repeatedly revised entries (details in Appendix~\ref{app:evidence_audit}).
We show examples of both successful and failure cases in Figures~\ref{fig:audit_update} and~\ref{fig:audit_examples}, respectively.

\subsection{Injected Error Detection}\label{sec:counterfactual}
Surfacing logical errors is an important aspect of reviewing, yet LLM reviewers often miss such errors~\citep{dycke2025automatic}.
To complement the overall quality evaluation, we test such ability of \methodname{} and baselines on 115 papers from the counterfactual dataset of \citet{dycke2025automatic}, each containing one deliberately injected error that perturbs a \textit{finding}, a \textit{result}, or a \textit{conclusion}; a detection counts as successful if it is mentioned in the weaknesses (cf. Appendix~\ref{app:counterfactual}).

\begin{table}[!htb]
\centering
\small
\definecolor{bestcell}{HTML}{D5F5E3}
\definecolor{secondcell}{HTML}{EBF5FB}
\providecommand{\best}[1]{\cellcolor{bestcell}\textbf{#1}}
\providecommand{\second}[1]{\cellcolor{secondcell}#1}
\setlength{\tabcolsep}{3pt}
\begin{tabular*}{\columnwidth}{@{}l@{\extracolsep{\fill}}*{4}{>{\centering\arraybackslash}p{0.9cm}}@{}}
\toprule
\textbf{Method} & \multicolumn{1}{c}{\textbf{Finding}} & \multicolumn{1}{c}{\textbf{Result}} & \multicolumn{1}{c}{\textbf{Concl.}} & \multicolumn{1}{c}{\textbf{All}} \\
\midrule
\multicolumn{5}{@{}l}{\textit{Prompt-based (Qwen3.5-397B)}} \\
\hspace{0.6em}AI-Scientist-v2 & .09 & \second{.26} & \best{.27} & \second{.21} \\
\hspace{0.6em}AgentReview     & .10 & .16 & .20 & .15 \\
\hspace{0.6em}TreeReview      & .10 & .20 & \second{.26} & .19 \\
\midrule
\multicolumn{5}{@{}l}{\textit{Fine-tuned (Qwen3-8B)}} \\
\hspace{0.6em}CycleReviewer   & .01 & .01 & .04 & .02 \\
\hspace{0.6em}DeepReview      & .03 & .02 & .09 & .05 \\
\hspace{0.6em}ReviewRL (Reimpl.) & \second{.12} & .16 & .22 & .17 \\
\hspace{0.6em}\textbf{\methodname{} (ours)} & \best{.24} & \best{.29} & \best{.27} & \best{.27} \\
\bottomrule
\end{tabular*}
\caption{Injected error detection rate.}
\label{tab:counterfactual}
\end{table}
As shown in Table~\ref{tab:counterfactual}, \methodname{} achieves the highest detection rate (27\%) with an 8B backbone, outperforming AI-Scientist-v2 on a 397B backbone by 6 percentage points, and is the only method that stays balanced across all three perturbation types (24--29\%).
The SFT baselines fall at or below 5\%, suggesting that they struggle to surface cross-section inconsistencies.
Although \methodname{} achieves the strongest performance, the task remains challenging for all methods. 
To better understand the remaining challenges, we analyze failure cases and find that most failures stem from overlooked \textit{numerical inconsistencies} and \textit{incorrect causal-direction reasoning}, as well as errors that are \textit{identified in the review log but not reported} in the final review (details in Appendix~\ref{app:counterfactual_failures}).

\section{Analysis}
\subsection{Component and Training Ablation}\label{sec:ablation}
 \begin{table}[t]
  \centering
  \small
  \newcommand{\drop}[1]{{\color{red!75!black}\scriptsize$\downarrow$#1}}
  \setlength{\tabcolsep}{2.5pt}
  \begin{tabular*}{\columnwidth}{@{\extracolsep{\fill}}lccccc r@{\hskip 2pt}l@{}}
  \toprule
  \textbf{Variant} & \textbf{Act.} & \textbf{Grd.} & \textbf{TD} & \textbf{Ver.} & \textbf{SA} & \multicolumn{2}{c}{\textbf{Avg.}} \\
  \midrule
  \textbf{\methodname{} (full)} & \textbf{.46} & \textbf{.64} & \textbf{.48} & \textbf{.40} & \textbf{.88} & \textbf{.57} & \\
  \midrule
  \multicolumn{8}{@{}l}{\textit{Architecture}} \\
  \hspace{0.6em}w/o Review Log & .31 & .26 & .24 & .34 & .59 & .35 & \drop{.22} \\
  \hspace{0.6em}w/o MDP & .30 & .52 & .35 & .39 & .66 & .44 & \drop{.13} \\
  \midrule
  \multicolumn{8}{@{}l}{\textit{Training}} \\
  \hspace{0.6em}SFT only & .12 & .23 & .15 & .14 & .31 & .19 & \drop{.38} \\
  \hspace{0.6em}No training & .12 & .19 & .13 & .08 & .13 & .13 & \drop{.44} \\
  \bottomrule
  \end{tabular*}
  \caption{Ablation of \methodname{} (Qwen3-8B). \drop{}: absolute drop in Avg.\ relative to the full model.}
  \label{tab:ablation}
\end{table}

To disentangle the contributions of different components and the training procedure, we perform controlled ablations.
Removing the review log leaves the investigation without an explicit evidence state, whereas removing the MDP keeps the evidence state but no longer lets it drive the investigation.

\paragraph{Within the architecture, the review log contributes most to performance.} As shown in Table~\ref{tab:ablation}, replacing the structured review log with free-form chain-of-thought leads to performance drops across all quality dimensions with the largest degradation in grounding (0.64 to 0.26).
Replacing the MDP with single-pass generation while retaining the review log also degrades performance across all dimensions.
These results indicate that the structured review log is the primary architectural contributor, with the MDP providing additional gains through iterative decision making.

\paragraph{Training improves execution validity and review quality in small models.}
The last two rows of Table~\ref{tab:ablation} isolate the contribution of each training stage.
SFT alone raises the average score from 0.13 to 0.19.
We find RL further improves both execution validity and review quality. 
It increases the rate of valid, format-compliant reviews from 29.2\% in the SFT model to 98.7\% and also improves the average score by 0.15 among successful reviews.
Format compliance and the associated review quality are reported in Appendix~\ref{app:validity}.

\subsection{\methodname{} as a Prompting Framework}\label{sec:prompting_strategy}

\begin{table}[t]
\centering
\small
\definecolor{bestcell}{HTML}{D5F5E3}
\definecolor{secondcell}{HTML}{EBF5FB}
\providecommand{\best}[1]{\cellcolor{bestcell}\textbf{#1}}
\providecommand{\second}[1]{\cellcolor{secondcell}#1}
\setlength{\tabcolsep}{2pt}
\begin{tabular*}{\columnwidth}{@{}l@{\extracolsep{\fill}}*{6}{>{\centering\arraybackslash}p{0.62cm}}@{}}
\toprule
\textbf{Framework} & \multicolumn{1}{c}{\textbf{Act.}} & \multicolumn{1}{c}{\textbf{Grd.}} & \multicolumn{1}{c}{\textbf{TD}} & \multicolumn{1}{c}{\textbf{Ver.}} & \multicolumn{1}{c}{\textbf{SA}} & \multicolumn{1}{c}{\textbf{Avg.}} \\
\midrule
AgentReview     & .28 & .46 & \second{.45} & .23 & .78 & .44 \\
AI-Scientist-v2 & .28 & .44 & \second{.45} & .25 & \best{.83} & .45 \\
TreeReview      & \second{.35} & \second{.66} & .42 & \second{.51} & \second{.80} & \second{.55} \\
\midrule
\textbf{\methodname{} (ours)} & \best{.46} & \best{.71} & \best{.50} & \best{.59} & .72 & \best{.60} \\
\bottomrule
\end{tabular*}
\caption{Training-free, on Qwen3.5-397B-A17B.}
\label{tab:prompting_framework}
\end{table}

To examine whether \methodname{} is effective without task-specific training, we apply the framework to Qwen3.5-397B-A17B through prompting and compare it with baselines using the same backbone.

As shown in Table~\ref{tab:prompting_framework}, \methodname{} achieves the highest average score (0.60), compared with 0.55 for the strongest baseline, TreeReview.
It leads on all four review-quality dimensions, although its score alignment is lower than that of the baselines.
These results show that \methodname{} is also effective when used as a prompting framework without any task-specific training.

\section{Conclusion}
We introduced \methodname{}, an evidence-driven review agent
that formulates investigation process as an MDP, with a structured review log that drives investigation and grounds the final review.
The trained 8B agent outperforms the compared baselines in average evaluation score and injected-error detection, while the method also remains effective as a prompting framework.
A human audit further shows that its reviews are largely supported by the evidence recorded in the review log.
These findings support evidence-driven investigation as a promising approach for LLM-assisted peer review, and potentially for other tasks requiring multi-step analytical reasoning over complex documents.

\section*{Limitations}
While \methodname{} achieves strong performance, there are several limitations that motivate future work to explore.

First, \methodname{} reasons over the submitted manuscript and therefore focuses on evidence, logical consistency within the paper itself.
It therefore does not evaluate review criteria that depend on external sources, such as novelty, completeness of related work, citation correctness, acceptance decision or other aspects requiring evidence beyond the manuscript.
This scope allows us to perform a fair comparison of intra-manuscript reasoning across all methods by controlling for differences in available evidence sources.
However, in practice, the framework itself can be naturally extended to incorporate different evidence sources by following the subagent-as-a-tool paradigm~\cite{ruan2026aorchestra}.
We present a proof-of-concept extension in Appendix~\ref{app:novelty_search} that equips \methodname{} with a novelty-checker subagent for literature search.

Second, the current implementation is text-only: the agent cannot directly inspect figures, which could include complementary evidence that is not accurately described in the text by the authors. Extending the agent with multimodal perception would allow it to verify visual claims (e.g., whether a reported trend matches a plotted curve).

Third, \methodname{} is trained and evaluated on AI conference papers (ICLR), as other fields currently lack sufficient publicly available, clean manuscript--review pairs. It is promising to adapt the approach to domains such as biomedicine or the social sciences once review data becomes available.

\section*{Acknowledgments}
This research work has been funded by the German Federal Ministry of Research, Technology, and Space and the Hessian Ministry of Higher Education, Research, Science, and the Arts within their joint support of the National Research Center for Applied Cybersecurity ATHENE. This work has been co-funded by the European Union (ERC, InterText, 101054961). Views and opinions expressed are, however, those of the author(s) only and do not necessarily reflect those of the European Union or the European Research Council. Neither the European Union nor the granting authority can be held responsible for them. We gratefully acknowledge support from the hessian.AI Service Center (funded by the Federal Ministry of Research, Technology and Space, BMFTR, grant no. 16IS22091) and the hessian.AI Innovation Lab (funded by the Hessian Ministry for Digital Strategy and Innovation, grant no. S-DIW04/0013/003). We express our sincere gratitude to Md Imbesat Hassan Rizvi, Serwar Basch, Sheng Lu, Qian Ruan, Fengyu Cai, and Frank Niu for their constructive feedback.

\section*{Ethical Considerations}
The development of \methodname{} carries several ethical considerations given its potential impact on the peer review process.
Its primary intended use is to help the authors of scientific papers identify potential issues and improve their work before submission and to provide supplementary reference to help reviewers identify possible issues, not as a final judgment. 
However, there are risks of misuse and unintended consequences that we discuss below.
An automated reviewing system could be misused to mass-produce superficial reviews or to game review assignment systems.
To mitigate this risk, we advocate for transparent disclosure whenever AI-generated reviews are used and recommend that venues establish clear policies governing their use.
Overreliance on automated reviews could also lead to reduced human oversight and potential erosion of review quality.
To address this, we emphasize that \methodname{} is designed to complement, not replace, human judgment, and we encourage users to critically evaluate its outputs rather than accepting them uncritically.
Additionally, our training and evaluation data consist of publicly available ICLR submissions and reviews from OpenReview. We use this data solely for research purposes and in accordance with its public availability. No private or confidential review data is used.  
\bibliography{custom}

\appendix
\clearpage
\section*{Appendix Contents}
\begingroup
\small
\setcounter{tocdepth}{-1}
\makeatletter
\@starttoc{toc}
\makeatother
\endgroup
\setcounter{tocdepth}{2}
\addtocontents{toc}{\protect\setcounter{tocdepth}{2}}
\vspace{1em}

\renewcommand{\topfraction}{0.9}
\renewcommand{\dbltopfraction}{0.9}
\renewcommand{\bottomfraction}{0.6}
\renewcommand{\textfraction}{0.1}
\renewcommand{\floatpagefraction}{0.8}
\renewcommand{\dblfloatpagefraction}{0.8}
\setcounter{topnumber}{3}
\setcounter{dbltopnumber}{3}
\setcounter{totalnumber}{5}

\section{Method and Training Details}\label{app:method_details}
\subsection{Action Schema}\label{app:action_schema}

Table~\ref{tab:action_schema} lists the complete action space available to the \methodname\ agent. Each turn, the agent outputs a JSON object with two fields: \texttt{action} (exactly one environment action) and \texttt{memory\_operations} (a list of zero or more log actions).
Both fields are generated from the current state before the environment returns a new observation. Log operations therefore add entries, update claim or question status, or add review points using information already observed; content retrieved by the environment action becomes available for subsequent decisions and log updates.

\begin{table*}[!t]
\centering
\small
\newcommand{\act}[3]{\multicolumn{2}{@{}l}{\texttt{\textbf{#1}}\texttt{(#2)}\quad --- #3}}
\newcommand{\argn}[1]{\hspace{1.2em}\texttt{#1}}
\newcommand{\opt}{\textcolor{gray}{\textit{(optional)}}}
\newcommand{\grp}[1]{\multicolumn{2}{@{}l}{\textit{#1}}}
\begin{tabularx}{\textwidth}{@{}p{3.6cm}>{\raggedright\arraybackslash}X@{}}
\toprule
\textbf{Action / argument} & \textbf{Description} \\
\midrule
\grp{Environment actions} \\
\addlinespace[3pt]
\rowcolor{gray!8}\act{read\_section}{section\_name}{Retrieve a section of the paper.} \\
\argn{section\_name} & Name of the section to read, as listed in the paper index. \\
\addlinespace[3pt]
\rowcolor{gray!8}\act{look\_up}{query}{Search the paper for passages matching a query.} \\
\argn{query} & Keywords or a phrase to search for. \\
\addlinespace[3pt]
\rowcolor{gray!8}\act{finish}{}{End the episode; the review outline becomes the final review. Takes no arguments.} \\
\midrule
\grp{Log operations} \\
\addlinespace[3pt]
\rowcolor{gray!8}\act{log}{type, text, section, \ldots}{Add a new entry to the review log.} \\
\argn{type} & Entry type: \texttt{claim}, \texttt{question}, or \texttt{note}. \\
\argn{text} & Content of the entry, citing concrete paper elements (e.g., Eq.~3, Table~2). \\
\argn{section} & Section of the paper the entry is drawn from. \\
\argn{claim\_type} & Category of a claim, e.g., \texttt{empirical}. \opt \\
\argn{issues} & Issues identified for a claim. \opt \\
\argn{question\_type} & Category of a question, e.g., \texttt{methodology}. \opt \\
\argn{related\_claims} & IDs of claims a question refers to. \opt \\
\addlinespace[3pt]
\rowcolor{gray!8}\act{update}{entry\_id, status, \ldots}{Update the status of an existing claim or question.} \\
\argn{entry\_id} & ID of the entry to update, e.g., \texttt{C1} or \texttt{Q3}. \\
\argn{status} & New status: \texttt{supported}, \texttt{weak}, \texttt{invalid}, or \texttt{to\_be\_verified} for claims; \texttt{resolved}, \texttt{partially\_answered}, or \texttt{open} for questions. \\
\argn{reasoning} & Justification for a claim's new status. \\
\argn{answer} & Answer that resolves a question. \\
\argn{cross\_references} & IDs of related entries. \opt \\
\argn{answer\_sections} & Sections that support the answer. \opt \\
\addlinespace[3pt]
\rowcolor{gray!8}\act{outline}{section, content, tags}{Add one point to the review outline.} \\
\argn{section} & Part of the review: \texttt{summary}, \texttt{strengths}, \texttt{weaknesses}, \texttt{questions}, or \texttt{overall\_score}. \\
\argn{content} & Text of the point, or an integer from 1 to 10 for \texttt{overall\_score}. \\
\argn{tags} & IDs of the log entries that support the point, e.g., \texttt{C1}, \texttt{Q2}, \texttt{N3}. \\
\bottomrule
\end{tabularx}
\caption{Action schema of \methodname{}. Environment actions retrieve paper content; log operations update the review log.}
\label{tab:action_schema}
\end{table*}

The review log maintained by the agent has four components:
\begin{itemize}
  \item \textbf{Claims} (\texttt{C1}, \texttt{C2}, \ldots): Assertions extracted from the paper, each with a verification status and source section.
  \item \textbf{Questions} (\texttt{Q1}, \texttt{Q2}, \ldots): Points of uncertainty raised during reading, each with a resolution status.
  \item \textbf{Notes} (\texttt{N1}, \texttt{N2}, \ldots): The agent's intermediate observations, plans, or thoughts.
  \item \textbf{Review Outline}: The developing structured review (summary, strengths, weaknesses, questions, overall score), where each strength and weakness must reference evidence IDs.
\end{itemize}

Below is an example of a single-turn agent output illustrating the JSON format:

\begin{reviewquotetag}
{\scriptsize
\begin{verbatim}
{
 "memory_operations": [
  {"op": "log", "args": {"type": "claim",
    "text": "Model achieves 92%
    "section": "4.2", "claim_type": "empirical"}},
  {"op": "update", "args": {"entry_id": "Q1",
    "status": "resolved",
    "answer": "Ablation in Table 3 confirms
              contribution"}}
 ],
 "action": {"name": "read_section",
            "args": {"section_name": "5. Analysis"}}
}
\end{verbatim}
}
\end{reviewquotetag}

\subsection{System Prompt}\label{app:system_prompt}

Figure~\ref{fig:system_prompt} shows the complete system prompt used by the \methodname\ agent during both RL training and inference.

\begin{figure*}[!t]
\begin{promptbox}[unbreakable]{System prompt of the \methodname\ agent}
\begin{lstlisting}[style=promptlisting]
# Task
You are reviewing a scientific paper. Your objective is to produce an accurate, internally consistent, and evidence-based review with: summary, strengths, weaknesses, questions for authors, and an overall score (1-10). You maintain a review log which help track your analysis and reasoning process. Your final review output is based on this log, so keep it updated and organized.

# Action space
Each turn, output a JSON object with two fields:
{ "memory_operations": [...], "action": {...} }
"memory_operations" is a list of log operations to update your review log (can be empty []). "action" is exactly one paper action.

## Paper Actions
- read_section: {"name": "read_section", "args": {"section_name": "..."}}
- look_up: {"name": "look_up", "args": {"query": "..."}}
- finish: {"name": "finish", "args": {}}  (You MUST call finish before running out of turns or your review is discarded.)

## Log Operations
- log: Record a new entry. Always use "op": "log".
  Claim:    {"op":"log","args":{"type":"claim","text":"...","section":"2.1","claim_type":"empirical"}}
  Question: {"op":"log","args":{"type":"question","text":"...","section":"3","question_type":"methodology"}}
  Note:     {"op":"log","args":{"type":"note","text":"...","section":"4"}}
  "section" is required. For claims, optionally add "issues": [...]. For questions, optionally add "related_claims": [...].
  All entries must cite concrete paper elements (e.g., Eq 3, Table 2, Fig 4(b)) and include specific details.
- update: Update the status of an existing claim or question.
  {"op":"update","args":{"entry_id":"C1|Q1","status":"...","reasoning":"..."}}
  For claims (C*): status in [supported, weak, invalid, to_be_verified]. Optionally include "cross_references": [...]
  For questions (Q*): status in [resolved, partially_answered, open]. Include "answer": "...". Optionally "answer_sections": [...]
- outline: Add one entry to your review outline.
  {"op":"outline","args":{"section":"summary|strengths|weaknesses|questions|overall_score","content":"...","tags":[...]}}
  For overall_score, content MUST be an integer 1-10. Each point in strengths/weaknesses MUST be grounded in the records
  (claims, questions, notes), reflected by the tags (C1, Q2, N3, etc). Every weakness and strength must cover a distinct issue.

# Review Log
You maintain a review log as persistent memory across turns. It has four components:
- Claims: Authors' statements you extracted, each with a verification status.
- Questions: Points of uncertainty or suspicion, each with a resolution status.
- Notes: Your observations, plans or thoughts.
- Review Outline: Your final verdict -- only add when confident based on evidence.
Guidelines:
- Use `log` to record claims, questions, and notes. Use `update` to change status after gathering evidence.
- Use `outline` to build your review. When you call `finish`, the outline becomes your final output.
- Verify all logged claims and answer all open questions. Never repeat a look_up with the same query.
- If multiple look_ups return no matches, the topic may be absent from the paper -- this itself can be a weakness.
Output valid JSON only.
\end{lstlisting}
\end{promptbox}
\caption{System prompt of the \methodname\ agent, used during both RL training and inference.}
\label{fig:system_prompt}
\end{figure*}

\subsection{Hyperparameters and Reward Weights}\label{app:hyperparams}

Table~\ref{tab:hyperparams} lists the key hyperparameters used in training.

\begin{table}[!ht]
\centering
\small
\begin{tabular}{@{}lcc@{}}
\toprule
\textbf{Hyperparameter} & \textbf{Qwen3-8B} & \textbf{Llama3.1-8B} \\
\midrule
\multicolumn{3}{@{}l}{\textit{SFT}} \\
\hspace{0.8em}Learning rate & $1 \times 10^{-5}$ & $1 \times 10^{-5}$ \\
\hspace{0.8em}Epochs & 3 & 3 \\
\hspace{0.8em}Global batch size & 64 & 64 \\
\hspace{0.8em}LR schedule & \multicolumn{2}{c}{Cosine (5\% warmup)} \\
\midrule
\multicolumn{3}{@{}l}{\textit{GRPO RL}} \\
\hspace{0.8em}Actor learning rate & $1 \times 10^{-6}$ & $5 \times 10^{-7}$ \\
\hspace{0.8em}KL coefficient & 0.01 & 0.001 \\
\hspace{0.8em}Train batch size & 16 & 16 \\
\hspace{0.8em}Rollouts per prompt ($n$) & 8 & 8 \\
\hspace{0.8em}Max episode steps & 30 & 30 \\
\midrule
\multicolumn{3}{@{}l}{\textit{Hardware}} \\
\hspace{0.8em}GPUs & \multicolumn{2}{c}{8$\times$A100 (80\,GiB)} \\
\bottomrule
\end{tabular}
\caption{Training hyperparameters for SFT and GRPO RL stages for \methodname{}.}
\label{tab:hyperparams}
\end{table}

\paragraph{Reward Weights.}
Training follows a two-phase curriculum. In Phase~1, only deterministic, rule-based rewards are active: step-level validity, review completeness, and score alignment.
Phase~2 retains all Phase~1 rewards with adjusted score alignment weight (1.0) and additionally introduces the LLM-judge-based content quality reward (weight 2.0), which combines technical depth and grounding (\S\ref{sec:reward}).

We simply assign weight 2.0 to the primary objective of each phase, while assigning weight 1.0 to the other active rewards.
Table~\ref{tab:reward_weighting} compares this scheme with equal weighting for the trained Qwen3-8B model.
The phase-specific weights yield a higher overall score (0.57 vs.\ 0.52).

\begin{table}[!ht]
\centering
\small
\setlength{\tabcolsep}{3.5pt}
\begin{tabular}{@{}lcccccc@{}}
\toprule
\textbf{Weighting} & \textbf{Act.} & \textbf{Grd.} & \textbf{TD} & \textbf{Ver.} & \textbf{SA} & \textbf{Avg.} \\
\midrule
Equal & 0.40 & 0.55 & 0.41 & 0.37 & \textbf{0.89} & 0.52 \\
Phase-specific (ours) & \textbf{0.46} & \textbf{0.64} & \textbf{0.48} & \textbf{0.40} & 0.88 & \textbf{0.57} \\
\bottomrule
\end{tabular}
\caption{Phase-specific versus equal reward weighting for \methodname{} (Qwen3-8B). \textbf{Bold}: better result. Act.: Actionability; Grd.: Grounding; TD: Technical Depth; Ver.: Verifiability; SA: Score Alignment.}
\label{tab:reward_weighting}
\end{table}

\section{Dataset Details}\label{app:data}

\paragraph{Training Data.}
We curate 4{,}011 submissions from ICLR 2025 for training and validation.
Paper manuscripts, including appendices, are fetched from the arXiv repository and converted from HTML to parseable Markdown format.
Since ICLR allows authors to update their manuscripts during the rebuttal period, we carefully match each paper's initial submission with its corresponding reviews and initial scores, ensuring that the review text is aligned with the manuscript version it assessed rather than a revised version modified in response to reviewer feedback.
For each paper, we collect the full set of official reviews, including textual assessments (summary, strengths, weaknesses, questions), overall ratings on a 1--10 scale, and reviewer confidence scores.
We removed author names, affiliations, user identifiers, email addresses, and other unnecessary personally identifying information, retaining only the content required for the research objectives.
After filtering for version alignment and review completeness, we split the data 90\%/10\% into training and validation sets.

\paragraph{Evaluation Data.}
For evaluation, we sample 1{,}000 papers from ICLR 2026 submissions, ensuring temporal separation from the training set to prevent data leakage.

\paragraph{Data Distribution.}
Table~\ref{tab:data_stats} summarizes the statistics of each data split.

\begin{table}[!ht]
\centering
\small
\setlength{\tabcolsep}{3pt}
\begin{tabular}{@{}lcccc@{}}
\toprule
\textbf{Split} & \textbf{\# Papers} & \textbf{Avg Tokens} & \textbf{Avg Rating} & \textbf{Accept \%} \\
\midrule
Train & 3{,}610 & 14{,}820 & 5.2 & 32.4 \\
Validation & 401 & 14{,}650 & 5.1 & 31.9 \\
Test & 1{,}000 & 15{,}130 & 5.3 & 33.1 \\
\bottomrule
\end{tabular}
\caption{Dataset statistics across splits. Avg Tokens is the mean paper length. Avg Rating is the mean overall score from human reviewers. Accept \% is the proportion of accepted papers.}
\label{tab:data_stats}
\end{table}

\paragraph{SFT Trace Generation.}
To produce supervised fine-tuning data, we use a teacher model (i.e., Qwen3.5-397B-A17B) to reconstruct the review process that would naturally produce a given human review. For each paper, we select multiple human reviews that are sufficiently detailed (long review text) and whose self-reported confidence is $\geq 4$, increasing the diversity and quality of the resulting traces. The teacher receives the paper and the human review (summary, strengths, weaknesses, questions, and overall score). It then generates a multi-turn interaction trace: reading sections, logging claims, raising questions, taking notes, verifying evidence, and incrementally building the review outline that faithfully reflects how a thorough reviewer would engage with the paper. 
The human review serves as a minimum coverage floor: the reconstructed trace must cover at least all points from the reference review but may include additional findings. 
This procedure yields 31{,}312 step-level training instances from 1{,}485 unique papers, grounded in actual human judgments without requiring human annotators to produce step-by-step traces.

\clearpage
\section{Evaluation Details}\label{app:eval_details}

\subsection{Evaluation Rubrics}\label{app:eval}

Since no single standardized rubric exists for review-quality evaluation, we construct a four-dimensional rubric grounded in prior work: we build on the utility framework of \citet{sadallah2025good} and integrate rubric designs from DeepReview~\citep{zhu2025deepreview}, ScholarPeer~\citep{goyal2026scholarpeer}, and CycleReviewer~\citep{weng2025cycleresearcher}.
Table~\ref{tab:rubrics} defines the score levels of each dimension.

\begin{table*}[!t]
\centering
\small
\renewcommand{\arraystretch}{1.15}
\begin{tabularx}{\textwidth}{@{}c*{4}{>{\raggedright\arraybackslash}X}@{}}
\toprule
& \textbf{Actionability} & \textbf{Grounding} & \textbf{Technical Depth} & \textbf{Verifiability} \\
& \textit{\footnotesize Is the action explicit or implicit, concrete or vague?}
& \textit{\footnotesize Is the addressed part identifiable, and the issue specified?}
& \textit{\footnotesize Does it engage with technical content and explain why it matters?}
& \textit{\footnotesize Is the claim supported by evidence, reasoning, or references?} \\
\midrule
\textbf{5} & Explicit action with concrete implementation details & Fully grounded and specific: names the part and the issue & Technical and reasoned: engages with methods, algorithms, or proofs and explains why the issue is a problem & Fully verifiable: thoroughly supported by explicit evidence, precise reasoning, or references \\
\addlinespace[2pt]
\textbf{4} & Implicit action with concrete guidance & Fully grounded but under-specific: names the part, not the issue & Technical but unreasoned: engages with technical content without explaining consequences & Mostly verifiable: well supported with minor gaps \\
\addlinespace[2pt]
\textbf{3} & Explicit action, vague on execution & Weakly grounded but specific: ambiguous part, clear issue & Non-technical but reasoned: explains why the gap matters & Somewhat verifiable: some justification, missing key elements \\
\addlinespace[2pt]
\textbf{2} & Implicit and vague action & Weakly grounded and not specific & Non-technical and unreasoned & Borderline verifiable: vague or insufficient support \\
\addlinespace[2pt]
\textbf{1} & No meaningful improvement guidance & Not grounded & No substance: surface observation only & Unverifiable: no supporting evidence or reasoning \\
\addlinespace[2pt]
\textbf{X} & --- & --- & --- & No claim (mapped to 0) \\
\bottomrule
\end{tabularx}
\caption{Evaluation rubrics for review quality. Scores range from 1 to 5 and are normalized to $[0,1]$.}
\label{tab:rubrics}
\end{table*}

We use three independent judges (GPT-5.4 nano, DeepSeek-V4 flash, and RevUtil) and average their scores. Per-judge results are reported in Appendix~\ref{app:per_judge}.

\subsection{Injected Error Detection Setup}\label{app:counterfactual}

\paragraph{Dataset.}
We use the counterfactual dataset introduced by \citet{dycke2025automatic}, which contains 138 papers from six conference iterations: ACL 2023, ACL 2024, EMNLP 2023, EMNLP 2024, NeurIPS 2024, and ICLR 2025.
Our evaluation excludes papers from ICLR 2025 to prevent data overlap and uses the remaining 115 papers.
Each paper has an original version and three counterfactual versions, each with one deliberately injected logical error of a distinct type, giving 115 papers per perturbation type.

\paragraph{Perturbation Types.}
Three types of errors are injected:
\begin{enumerate}
  \item \textbf{Conclusion perturbation}: Alters a conclusion to misalign with its underlying result.
  \item \textbf{Finding perturbation}: Exaggerates the claim of a finding beyond what the evidence supports.
  \item \textbf{Result perturbation}: Changes a result to contradict the conclusion it originally supported.
\end{enumerate}
Each counterfactual paper includes metadata specifying the modification type, the modified claim, and the logical relationship explaining why the injected claim is incorrect.

\paragraph{Detection Judgment.}
Each LLM review method generates a review of the counterfactual paper. We then use an LLM judge (GPT-5.4 nano) to determine whether any weakness in the generated review identifies the injected error. The judge receives:
\begin{itemize}
  \item The injected error description (type, modified claim, and why it is wrong).
  \item The list of weaknesses from the generated review.
\end{itemize}
The judge outputs a JSON with four fields: \texttt{detected} (true/false), \texttt{confidence} (high/medium/low), \texttt{matching\_weakness\_index}, and \texttt{reasoning}.
A detection is counted as successful if the judge returns \texttt{detected: true} with \texttt{confidence: high}.
We validate the adjudicator against human annotations on a randomly sampled subset (50 instances) of its decisions.
Agreement between human annotations and the LLM judge is Cohen's $\kappa=0.73$, and the judge's detection precision is 0.88.
False positives arise when a weakness is semantically related to an injected claim but does not identify the specific logical error. For example, a generic request to discuss generalizability can be mistaken for identifying that the reported experiments do not support a claim of cross-domain generalization.

\section{Human Evaluation}\label{app:human_eval}

\paragraph{Evaluators.}
We recruit 5 expert evaluators who have served as reviewers for top-tier AI conferences (e.g., NeurIPS, ICLR, ACL, EMNLP).

\paragraph{Evaluation Protocol.}
Human evaluation uses pairwise comparison of reviews. For each paper, evaluators compare pairs of reviews from different systems. Each comparison presents the paper and two anonymized reviews (labeled ``Review A'' and ``Review B''), and evaluators select which review provides higher-quality feedback.

\paragraph{Systems Compared.}
Five systems are included: AgentReview, AI-Scientist-v2, CycleReviewer, DeepReview, and \methodname{}. Systems are assigned anonymous identifiers (\texttt{system\_P} through \texttt{system\_T}) with a seeded random permutation; evaluators never see real system names. For each paper, 5 pairwise comparisons are generated from a rotating cycle design that ensures all 10 possible system pairs are covered in aggregate, with each system appearing twice per paper. The A/B presentation order is randomized with a coin flip per comparison.

\paragraph{Supplementary Comparisons with TreeReview and ReviewRL.}
TreeReview and ReviewRL (Reimpl.) were added after the initial evaluation and compared against \methodname{} in a supplementary round with the same evaluators, papers, and protocol. The Bradley--Terry model in Table~\ref{tab:bt_scores} is fit jointly on both rounds.

\paragraph{Paper Selection and Overlap.}
50 papers are randomly sampled from the test set. Each evaluator reviews 30 papers (150 pairwise comparisons). Approximately 20\% of each evaluator's papers are shared across all evaluators to enable inter-annotator agreement measurement.

\subsection{Annotator Guidelines}\label{app:annotator_guide}

The complete guidelines provided to human evaluators are reproduced below.

\begin{promptbox}{Annotator guidelines for pairwise review-quality evaluation}
\begin{lstlisting}[style=promptlisting]
# Annotator Guidelines: Review Quality Pairwise Evaluation

## Overview
You will compare pairs of peer reviews for the same academic paper across 4 quality dimensions. For each comparison, you see:
- Paper Info: The paper title, abstract, and a link to the full paper on arXiv
- Review A and Review B: Two anonymized reviews
For each dimension, judge which review is better, or whether they are tied (A wins / B wins / Tie).

## Dimension 1: Actionability
Which review provides more actionable feedback?
Consider: Are suggestions explicit or implicit? Concrete or vague? Does the review tell authors *how* to improve?
  (5) Highly Actionable: Explicit actions with concrete implementation details
  (4) Mostly Actionable: Implicit actions but concrete execution details
  (3) Somewhat Actionable: Explicit actions but vague on execution
  (2) Borderline Actionable: Implicit and vague actions
  (1) Unactionable: Pure observations without suggestions

## Dimension 2: Grounding
Which review better grounds its critiques in specific parts of the paper?
Consider: Does it reference specific sections, tables, figures, equations? Or make generic claims?
  (5) Fully Grounded and Specific: References sections/tables/figures AND specifies what is wrong
  (4) Fully Grounded but Under-Specific: References location but vague about the issue
  (3) Weakly Grounded and Specific: No structural reference but clear about the issue
  (2) Weakly Grounded and Not Specific: Neither referenced nor specific
  (1) Not Grounded: No identifiable paper section addressed

## Dimension 3: Verifiability
Which review better supports its claims with justification?
Consider: Does the reviewer provide logical reasoning, common knowledge, or external references?
  (5) Fully Verifiable: Claims thoroughly supported by reasoning or references
  (4) Mostly Verifiable: Well-supported but minor gaps
  (3) Somewhat Verifiable: Some justification but lacks key elements
  (2) Borderline Verifiable: Vague or insufficient support
  (1) Unverifiable: No supporting evidence or justification

## Dimension 4: Technical Depth
Which review demonstrates stronger technical engagement and reasoning?
Consider: Does it address specific technical content and explain why the issue matters?
  (5) Technical and reasoned: Engages with methodology, algorithms, or proofs and explains why the issue is problematic
  (4) Technical but unreasoned: Engages with technical content without explaining consequences
  (3) Non-technical but reasoned: Provides reasoning about why a gap matters without engaging with specific technical content
  (2) Non-technical and unreasoned
  (1) No substance: Pure surface observation

## Important Notes
- Evaluate each dimension independently.
- Ignore review length -- a shorter, focused review can be better than a longer, vague one.
- Do not try to identify the systems; reviews are anonymized.
- Be consistent. Apply the same standards across all comparisons.
- When in doubt, re-read the paper via the provided arXiv link.
\end{lstlisting}
\end{promptbox}

\subsection{Bradley--Terry Analysis}\label{app:human_eval_bt}

Based on the evaluated data, we fit a Bradley--Terry (BT) model to derive proper strength estimates. Ties are split as 0.5 wins for each side. Scores are reported on an Elo-like scale (400 points $\approx$ 10$\times$ strength ratio), anchored at 1000.

\begin{table}[!ht]
\centering
\small
\setlength{\tabcolsep}{2.5pt}
\begin{tabular}{@{}lcccc@{}}
\toprule
\textbf{System} & \textbf{Act.} & \textbf{Grd.} & \textbf{TD} & \textbf{Ver.} \\
\midrule
\methodname{} & \textbf{1127} & \textbf{1272} & \textbf{1208} & 1149 \\
[-1pt] & {\tiny\color{gray}[1083, 1178]} & {\tiny\color{gray}[1214, 1346]} & {\tiny\color{gray}[1155, 1274]} & {\tiny\color{gray}[1102, 1205]} \\[1.5pt]
TreeReview & 1015 & 1118 & 1010 & \textbf{1158} \\
[-1pt] & {\tiny\color{gray}[915, 1115]} & {\tiny\color{gray}[1016, 1220]} & {\tiny\color{gray}[891, 1129]} & {\tiny\color{gray}[1055, 1261]} \\[1.5pt]
AI-Scientist-v2 & 964 & 1011 & 1032 & 1014 \\
[-1pt] & {\tiny\color{gray}[920, 1006]} & {\tiny\color{gray}[969, 1053]} & {\tiny\color{gray}[991, 1075]} & {\tiny\color{gray}[971, 1056]} \\[1.5pt]
AgentReview & 971 & 945 & 982 & 952 \\
[-1pt] & {\tiny\color{gray}[930, 1012]} & {\tiny\color{gray}[899, 987]} & {\tiny\color{gray}[938, 1025]} & {\tiny\color{gray}[911, 993]} \\[1.5pt]
DeepReview & 1015 & 893 & 951 & 940 \\
[-1pt] & {\tiny\color{gray}[973, 1059]} & {\tiny\color{gray}[841, 938]} & {\tiny\color{gray}[901, 999]} & {\tiny\color{gray}[893, 985]} \\[1.5pt]
ReviewRL (Reimpl.) & 998 & 1042 & 956 & 1003 \\
[-1pt] & {\tiny\color{gray}[895, 1101]} & {\tiny\color{gray}[924, 1160]} & {\tiny\color{gray}[833, 1078]} & {\tiny\color{gray}[899, 1108]} \\[1.5pt]
CycleReviewer & 908 & 760 & 817 & 786 \\
[-1pt] & {\tiny\color{gray}[861, 951]} & {\tiny\color{gray}[711, 803]} & {\tiny\color{gray}[764, 865]} & {\tiny\color{gray}[736, 828]} \\[1.5pt]
\bottomrule
\end{tabular}
\caption{Bradley--Terry Elo scores, with 95\% bootstrap confidence intervals (2{,}000 resamples) in gray. \textbf{Bold}: best per dimension. Act.: Actionability, Grd.: Grounding, TD: Technical Depth, Ver.: Verifiability.}
\label{tab:bt_scores}
\end{table}

Table~\ref{tab:bt_scores} shows the full ranking. \methodname{} achieves the highest BT score on Actionability, Grounding, and Technical Depth, while TreeReview ranks first on Verifiability by a small margin, consistent with the pairwise results in Figure~\ref{fig:human_eval}.

\subsection{Inter-Annotator Agreement}\label{app:human_eval_iaa}

We measure inter-annotator agreement on the 20\% overlap set using three chance-corrected metrics: Krippendorff's $\alpha$, Fleiss' $\kappa$, and average pairwise quadratic-weighted Cohen's $\kappa^{2}$.

\begin{table}[!ht]
\centering
\small
\begin{tabular}{@{}lccc@{}}
\toprule
\textbf{Dimension} & \textbf{Kripp.\ $\alpha$} & \textbf{Fleiss' $\kappa$} & \textbf{Cohen's $\kappa^{2}$} \\
\midrule
Actionability    & .44 & .52 & .46 \\
Grounding        & .70 & .68 & .50 \\
Technical Depth  & .56 & .54 & .52 \\
Verifiability    & .55 & .43 & .66 \\
\bottomrule
\end{tabular}
\caption{Inter-annotator agreement across dimensions on the overlap set.}
\label{tab:iaa}
\end{table}

\subsection{Evidence Trail Audit}\label{app:evidence_audit}

We audit two links behind each review point: whether it is supported by its cited log entries (\emph{review-to-log support}), and whether those entries are faithful to the manuscript (\emph{log-to-paper faithfulness}).
We randomly sample one review from each of 30 papers and two strengths and two weaknesses from each review, giving 120 review points that cite 172 log entries (126 claims, 34 questions, 12 notes).

\paragraph{Annotators.}
Two expert reviewers annotate with access to the full manuscript, the review, and the review log.
Each paper is labeled by one annotator; on 6 papers labeled by both, agreement is 95\% (Cohen's $\kappa=0.66$, lowered by the skewed label distribution).

\paragraph{Labels.}
A review point is \textbf{Supported} if its cited entries together provide direct evidence for its assertions, and \textbf{Unsupported} otherwise (e.g., the entries are only on-topic, unrelated, or contradictory).
A cited entry is \textbf{Faithful} if it accurately reflects the manuscript, and \textbf{Unfaithful} otherwise.
A review point satisfies both criteria if it is Supported and all its cited entries are Faithful.

\begin{table}[!ht]
\centering
\small
\begin{tabular}{@{}lrrr@{}}
\toprule
\textbf{Entry type} & \textbf{Total} & \textbf{Faithful} & \textbf{Unfaithful} \\
\midrule
Claim    & 126 & 112 (88.9\%) & 14 (11.1\%) \\
Question &  34 &  27 (79.4\%) &  7 (20.6\%) \\
Note     &  12 &  11 (91.7\%) &  1 (8.3\%) \\
\midrule
\textbf{All} & \textbf{172} & \textbf{150 (87.2\%)} & \textbf{22 (12.8\%)} \\
\bottomrule
\end{tabular}
\caption{Log-to-paper faithfulness of the 172 cited entries by entry type, as number of entries (percentage).}
\label{tab:audit_faithfulness}
\end{table}

\paragraph{Results.}
98.3\% of review points are Supported, 87.2\% of cited entries are Faithful (Table~\ref{tab:audit_faithfulness}), and 84.2\% of review points satisfy both criteria.

\paragraph{Update operation.}
\textbf{Frequency.} The agent actively revises its evidence rather than recording it once: in our manuscript-faithfulness audit, 86.9\% of cited claims and questions are updated at least once, and 22.5\% more than once (up to five rounds) before reaching a final status.
\textbf{Accuracy.} We manually check 30 updated entries against the source manuscripts. The sample is stratified to include all 17 cases where an update reversed an earlier verdict (e.g., from \texttt{supported} to \texttt{invalid}), and thus over-represents difficult cases. Even so, 26 entries (86.7\%) reach a final status faithful to the manuscript, and 4 are unfaithful.
\textbf{Where errors occur.} The unfaithful updates include one that cites a nonexistent table to mark a question \texttt{resolved}, and another that assigns a status contradicting its own stated reasoning.
Overall, \textit{update} largely finalizes under-verified entries correctly once new evidence appears; the few remaining errors arise when an update is not backed by the manuscript or is inconsistent with its own reasoning, and occur only in entries that are revised multiple times or have their verdict reversed.

\paragraph{Audit examples.}\label{app:audit_failures}
Figure~\ref{fig:audit_update} shows review points that satisfy both criteria: panel~(a) shows an entry refined by an \textit{update} operation after new evidence is read, and panel~(b) shows a weakness fully supported by a single faithful entry.
Figure~\ref{fig:audit_examples} shows \emph{unfaithful log entries}: in both cases the review point is Supported by its cited entry, but the entry misrepresents the manuscript.

\begin{figure*}[t]
\centering
\begin{tcolorbox}[auditstyle, fonttitle=\small, title={\textbf{(a) From an open question to an evidence-based weakness} --- OptPipe, MILP solution quality, \S5.2}]
\footnotesize
\raggedright
\begin{minipage}[t]{0.30\linewidth}
\textbf{1. Manuscript evidence}\par\smallskip
\S5.2 specifies solver time limits of 300\,s for 4/8 GPUs and \textbf{1000\,s for 16 GPUs}, selecting the \textit{``\textbf{best feasible solution within the limit.}''}\par\smallskip
For larger cases, the solver often reaches the time limit. The authors state that their heuristics obtain \textit{``high-quality solutions within the limit, \textbf{even if they are not optimal.}''}\par\smallskip
However, \textcolor{green!40!black}{\textbf{no numerical optimality gap is reported}} to quantify how close these feasible solutions are to the optimum.
\end{minipage}\hfill
\begin{minipage}[t]{0.32\linewidth}
\textbf{2. Q3 evolves through an update}\par\smallskip
\textbf{Step 7 --- Initial question:}\par
``For the largest cases (16 GPUs) where the solver hits the time limit, how close are the obtained solutions to the theoretical optimum? \textcolor{green!40!black}{\textbf{Is there an optimality gap reported}}\ldots{}?''\par\smallskip
\centering$\Downarrow$\par
\textcolor{green!40!black}{\textbf{\texttt{update} grounded in \S5.2}}\par
\raggedright\textbf{Step 8 --- After:} \textcolor{green!40!black}{\textbf{\texttt{partially\_answered}}}\par
``However, the paper \textcolor{green!40!black}{\textbf{does not report the specific optimality gap}}\ldots{} Without this metric, it is unclear if the returned schedules are near-optimal or significantly suboptimal, though the empirical performance suggests they are practically useful.''
\end{minipage}\hfill
\begin{minipage}[t]{0.30\linewidth}
\textbf{3. Final review point}\par\smallskip
\textbf{Weakness 1}\par
\textit{Outline evidence tags: \texttt{C14}, \texttt{Q3}.}\par\smallskip
``The evaluation \textcolor{green!40!black}{\textbf{lacks reporting of the MILP optimality gap}} for large-scale experiments (e.g., 16 GPUs) where the solver hits the 1000s time limit. While the authors state they use the `best feasible solution,' without the gap metric, it is impossible to assess how close these schedules are to the theoretical optimum\ldots{}''\par\smallskip
The updated question becomes a concrete reporting weakness: empirical usefulness is demonstrated, but \textcolor{green!40!black}{\textbf{proximity to the optimum remains unquantified}}.
\end{minipage}
\par\medskip\hrule\smallskip
\textbf{Manuscript $\rightarrow$ log:} \textcolor{green!40!black}{\textbf{Faithful}} --- the update retains the time-limited solver setting while identifying the missing optimality-gap measurement.\par\smallskip
\textbf{Log $\rightarrow$ review:} \textcolor{green!40!black}{\textbf{Fully Supported}} --- the final weakness carries forward Q3's unresolved concern about solution quality. Its outline cites Q3 and C14 (C14 is omitted from this excerpt).\par\smallskip
\textbf{Joint criterion: Pass} --- the updated entry is faithful to the manuscript and supports the weakness about missing optimality-gap reporting.
\end{tcolorbox}
\begin{tcolorbox}[auditstyle, fonttitle=\small, title={\textbf{(b) Deep technical weakness with a single-entry chain} --- Diffusion LLM decoding, \S5.1}]
\footnotesize
\begin{minipage}[t]{0.30\linewidth}
\textbf{1. Manuscript evidence}\par\smallskip
At 2048-token sequences, \textcolor{green!40!black}{\textbf{performance degrades due to a distribution shift between training with the full suffix and inference with the suffix dropped}}. Section~5.1 reports that \textcolor{green!40!black}{\textbf{fine-tuning with a revised objective using stochastic dropout masks (Eq.~13) mitigates the degradation}}.
\end{minipage}\hfill
\begin{minipage}[t]{0.32\linewidth}
\textbf{2. Log entry C15}\par\smallskip
``DPad experiences \textcolor{green!40!black}{\textbf{performance degradation on very long sequences (2048 tokens) due to a distributional shift between training (full suffix) and inference (dropped suffix)}}, which can be mitigated by \textcolor{green!40!black}{\textbf{fine-tuning with a revised objective incorporating stochastic dropout masks}}.''
\end{minipage}\hfill
\begin{minipage}[t]{0.30\linewidth}
\textbf{3. Final review point}\par\smallskip
\textbf{Weakness [C15]:}\par
``At 2048-token sequences, the method suffers \textcolor{green!40!black}{\textbf{performance degradation due to a train--inference distribution shift between full and dropped suffixes}}. Although \textcolor{green!40!black}{\textbf{fine-tuning with stochastic dropout masks mitigates this issue}}, it adds a training step for the long-context setting (\S5.1, Eq.~13).''
\end{minipage}
\par\medskip\hrule\smallskip
\textbf{Manuscript $\rightarrow$ log:} \textcolor{green!40!black}{\textbf{Faithful}} --- C15 accurately records the degradation, its stated cause, and the reported mitigation in \S5.1.\par\smallskip
\textbf{Log $\rightarrow$ review:} \textcolor{green!40!black}{\textbf{Fully Supported}} --- C15 supports the degradation, the train--inference mismatch, and the fine-tuning mitigation described in the weakness.\par\smallskip
\textbf{Joint criterion: Pass} --- the single cited entry faithfully supports the review point.
\end{tcolorbox}
\caption{Audit examples satisfying both criteria. Panel (a) traces an optimality question through an evidence-based \texttt{update} to a concrete weakness about missing optimality-gap reporting. Panel (b) shows a weakness whose single cited entry (C15) supports the reported degradation, its cause, and the mitigation.}
\label{fig:audit_update}
\end{figure*}

\begin{figure*}[t]
\centering
\begin{tcolorbox}[auditstyle, fonttitle=\small, title={\textbf{(c) Value read from the wrong table row} --- PDE imputation, Appendix G.2, Table 6}]
\footnotesize
\begin{minipage}[t]{0.30\linewidth}
\textbf{1. Manuscript evidence}\par\smallskip
Table 6 (Navier--Stokes, 60\% column):\par\smallskip
70\%/70\% ratio: \textbf{0.5336}\par
90\%/90\% ratio: \textbf{0.5251}\par
100\%/100\% (MissDiff): 0.7023
\end{minipage}\hfill
\begin{minipage}[t]{0.32\linewidth}
\textbf{2. Log entry C16}\par
\texttt{to\_be\_verified}\par\smallskip
``Appendix G.2 Table 6 validates the theoretical trade-off\ldots{} (\textbf{70\%/70\%}) outperform extremes\ldots{} (e.g., 0.7023 vs \textcolor{red!70!black}{\textbf{0.5251}} on Navier-Stokes 60\%)\ldots{}''\par\smallskip
0.5251 belongs to the 90\%/90\% row; the 70\%/70\% value in this column is 0.5336.
\end{minipage}\hfill
\begin{minipage}[t]{0.30\linewidth}
\textbf{3. Final review point}\par\smallskip
\textbf{Strength [C14, C16, Q4]:}\par
``\ldots{} Table 4 confirms the critical necessity of the ensemble sampling strategy (error increases $>$12$\times$ without it), while \textcolor{red!70!black}{\textbf{Table 6 proves that the context-query partitioning mechanism itself drives performance gains over the MissDiff baseline}} (which corresponds to 100\% ratios).''
\end{minipage}
\par\medskip\hrule\smallskip
\textbf{Manuscript $\rightarrow$ log:} \textcolor{red!70!black}{\textbf{Unfaithful}} --- C16 assigns the 90\%/90\% value (0.5251) to the 70\%/70\% condition, misattributing a real number to the wrong experimental setting.\par\smallskip
\textbf{Log $\rightarrow$ review:} \textcolor{green!40!black}{\textbf{Supported}} --- the cited log supports the highlighted claim about the partitioning mechanism's gains over MissDiff, despite the numerical misattribution in C16.
\par\smallskip
\textbf{Joint criterion: Fail} --- the review point cites an entry that misattributes manuscript evidence, even though the direction of the numerical comparison is unchanged.
\end{tcolorbox}

\begin{tcolorbox}[auditstyle, fonttitle=\small, title={\textbf{(d) Unsupported attribution of an explanation} --- MoE pruning, \S4.2}]
\footnotesize
\begin{minipage}[t]{0.30\linewidth}
\textbf{1. Manuscript evidence}\par\smallskip
At compression ratio 0.9: ``about 55\% of its baseline accuracy'' with ``a 38\% reduction in FLOPs.''\par\smallskip
At low compression: ``the accuracy curve remains nearly flat, revealing substantial redundancy among micro-experts.''
\end{minipage}\hfill
\begin{minipage}[t]{0.32\linewidth}
\textbf{2. Log entry Q3 (resolved)}\par\smallskip
Answer: ``Section 4.2 and Figure 2 clarify that\ldots{} \textcolor{red!70!black}{\textbf{Attention layers (which are not pruned) dominate the remaining cost}}.''\par\smallskip
The entry also attributes the limited FLOPs savings to an attention bottleneck using the phrase ``redundancy among micro-experts.''
\end{minipage}\hfill
\begin{minipage}[t]{0.30\linewidth}
\textbf{3. Final review point}\par\smallskip
\textbf{Weakness [C19, Q3]:}\par
``Figure 2 analysis reveals a disconnect between parameter reduction and FLOPs savings: removing 90\% of parameters (0.9 compression) yields only 38\% FLOPs reduction (C19). This suggests the method primarily targets parameter memory footprint rather than inference latency, as \textcolor{red!70!black}{\textbf{Attention layers (unpruned) dominate the computational cost in high-sparsity regimes.}} The paper could better clarify this trade-off for users prioritizing speed over memory.
\end{minipage}
\par\medskip\hrule\smallskip
\textbf{Manuscript $\rightarrow$ log:} \textcolor{red!70!black}{\textbf{Unfaithful}} --- Q3 attributes an attention-bottleneck explanation to the manuscript, but the cited passage reports accuracy and FLOPs trade-offs without establishing that explanation.\par\smallskip
\textbf{Log $\rightarrow$ review:} \textcolor{green!40!black}{\textbf{Supported}} --- the cited entries support the review's criticism, including the attention-bottleneck explanation recorded in Q3.
\par\smallskip
\textbf{Joint criterion: Fail} --- the judgment relies on an explanation not established by the cited manuscript evidence.
\end{tcolorbox}

\caption{Unfaithful log entries from the evidence-trail audit: (c) a table misreading and (d) an unsupported interpretation. In both cases, the review point is Supported by its cited entry, but the entry misrepresents the manuscript.}
\label{fig:audit_examples}
\end{figure*}

\section{Additional Results}\label{app:additional_results}

\subsection{Per-Judge Evaluation Results}\label{app:per_judge}

Tables~\ref{tab:gpt54_results}--\ref{tab:utility_results} present the full evaluation results for each of the three judges used in our evaluation: GPT-5.4 nano (Table~\ref{tab:gpt54_results}), DeepSeek-V4 flash (Table~\ref{tab:deepseekv4_results}), and the utility-based RevUtil judge (Table~\ref{tab:utility_results}). The main paper reports averages across all three judges.

\paragraph{Robustness to judge-specific score scales.}
Since the three judges differ in how strictly they score and how widely their scores spread, we standardize each judge's scores per dimension into z-scores across systems before averaging. As shown in Table~\ref{tab:judge_robustness}, the ranking is largely unchanged: \methodname{} (Qwen3-8B) remains first under both avg@4 and best@4, and the top five systems keep their order.

\begin{table}[!ht]
\centering
\small
\setlength{\tabcolsep}{4pt}
\resizebox{\columnwidth}{!}{%
\begin{tabular}{@{}llcccc@{}}
\toprule
& & \multicolumn{2}{c}{\textbf{avg@4}} & \multicolumn{2}{c}{\textbf{best@4}} \\
\cmidrule(lr){3-4} \cmidrule(lr){5-6}
\textbf{Method} & \textbf{Model} & Raw & $z$ & Raw & $z$ \\
\midrule
\methodname{} & Qwen3-8B & \textbf{1} & \textbf{1} & \textbf{1} & \textbf{1} \\
TreeReview & Qwen3.5-397B & 2 & 2 & 3 & 3 \\
\methodname{} & Llama3.1-8B & 3 & 3 & 2 & 2 \\
ReviewRL & Qwen3-8B & 4 & 4 & 4 & 4 \\
TreeReview & Gemini-3.1-FL & 4 & 5 & 5 & 5 \\
AI-Scientist-v2 & Qwen3.5-397B & 6 & 6 & 7 & 7 \\
AgentReview & Qwen3.5-397B & 7 & 7 & 7 & 8 \\
DeepReview & Llama3.1-8B & 8 & 8 & 6 & 6 \\
AgentReview & Gemini-3.1-FL & 8 & 9 & 9 & 9 \\
CycleReviewer & Llama3.1-8B & 10 & 10 & 12 & 11 \\
AI-Scientist-v2 & Gemini-3.1-FL & 10 & 11 & 10 & 12 \\
CycleReviewer & Qwen3-8B & 12 & 12 & 14 & 14 \\
DeepReview & Qwen3-8B & 13 & 13 & 12 & 10 \\
TreeReview & Qwen3-8B & 14 & 14 & 10 & 13 \\
AgentReview & Qwen3-8B & 15 & 15 & 15 & 16 \\
TreeReview & Llama3.1-8B & 16 & 16 & 15 & 15 \\
AI-Scientist-v2 & Qwen3-8B & 17 & 17 & 17 & 18 \\
AI-Scientist-v2 & Llama3.1-8B & 18 & 18 & 17 & 17 \\
AgentReview & Llama3.1-8B & 19 & 19 & 19 & 19 \\
\bottomrule
\end{tabular}%
}
\caption{Overall ranks of the systems in Table~\ref{tab:main_results} under raw-score averaging (Raw; ties share a rank) and within-judge standardization ($z$).}
\label{tab:judge_robustness}
\end{table}

\subsection{Results with Intermediate-Size Backbones}\label{app:mid_size}

Table~\ref{tab:mid_size} reports the prompt-based baselines with Qwen3-14B and Qwen3-32B backbones, evaluated with the same protocol as Table~\ref{tab:main_results}.

\subsection{Format Compliance and Review Quality}\label{app:validity}

Table~\ref{tab:validity} reports, for each ablation, the percentage of format-compliant reviews, the overall score across all reviews, and the score restricted to format-compliant reviews.

\begin{table}[!ht]
\centering
\small
\setlength{\tabcolsep}{6pt}
\resizebox{\columnwidth}{!}{%
\begin{tabular}{@{}lccc@{}}
\toprule
\textbf{Method} & \shortstack{\textbf{Format-compliant}\\\textbf{(\%)}} & \shortstack{\textbf{Overall}\\\textbf{score}} & \shortstack{\textbf{Score on}\\\textbf{compliant reviews}} \\
\midrule
\methodname{} & 98.7 & 0.57 & 0.58 \\
\midrule
w/o Review Log & 98.8 & 0.35 & 0.35 \\
w/o MDP & 99.5 & 0.44 & 0.45 \\
\midrule
SFT only & 29.2 & 0.19 & 0.43 \\
No training & 27.8 & 0.13 & 0.32 \\
\bottomrule
\end{tabular}}
\caption{Format compliance and review quality for the ablations in Table~\ref{tab:ablation}. Both scores average the five evaluation dimensions. The overall score includes all reviews, whereas the score on compliant reviews includes only format-compliant outputs.}
\label{tab:validity}
\end{table}

\subsection{Robustness to Paper Length}\label{app:paper_length}
\begin{figure*}[!t]
\centering
\includegraphics[width=\textwidth]{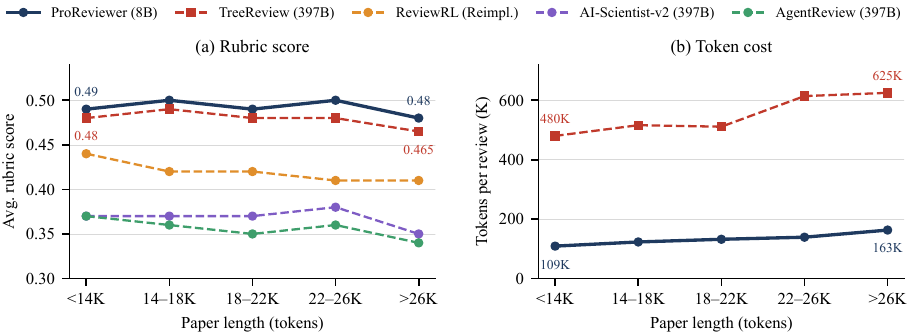}
\caption{Paper-length analysis. (a) Average rubric score (mean of the four quality dimensions) per length bin. (b) Total tokens (input and output) per review. 8B = Qwen3-8B; 397B = Qwen3.5-397B-A17B.}
\label{fig:length-sensitivity}
\end{figure*}

We hypothesize that \methodname{}'s iterative investigation strategy makes it robust to increasing paper length by selectively inspecting relevant parts of the paper, rather than processing the entire manuscript at each step.
To test this, we partition the test set into five bins by token count (Table~\ref{tab:length_bins}) and report the average rubric score (mean of the four quality dimensions) per bin.
As shown in Figure~\ref{fig:length-sensitivity}(a), \methodname{} (Qwen3-8B) remains essentially flat across bins, decreasing by only $1\%$ relative from the shortest to the longest papers.
TreeReview (Qwen3.5-397B-A17B) is similarly stable (a $3\%$ relative decrease), whereas the other baselines decline by $5.4\%$--$8.1\%$ relative.
Figure~\ref{fig:length-sensitivity}(b) shows the cost of this robustness.
TreeReview consumes about four times as many tokens per review as \methodname{}, rising from 480K to 625K on the longest papers.
This growth follows the size of its question tree, which expands from 85 nodes on average for papers under 14K tokens to 102 for papers of 22--26K tokens, consistent with the cost scaling noted in Section~\ref{sec:related}.
\methodname{} grows from 109K to 163K tokens, sublinearly in paper length: each step processes only a compact working context, and the number of steps stays nearly constant (19.9 to 21.4 on average; Table~\ref{tab:length_cost}).

\begin{table}[!ht]
\centering
\small
\setlength{\tabcolsep}{4pt}
\begin{tabular}{@{}lcccc@{}}
\toprule
\textbf{Bin} & \textbf{Steps} & \textbf{Input} & \textbf{Gen.} & \textbf{Total} \\
\midrule
0--14K  & 19.9{\scriptsize$\pm$6.3} & 103K & 6.1K & 109K \\
14--18K & 20.2{\scriptsize$\pm$6.1} & 116K & 6.5K & 123K \\
18--22K & 20.3{\scriptsize$\pm$6.1} & 126K & 6.8K & 132K \\
22--26K & 20.4{\scriptsize$\pm$6.3} & 132K & 6.9K & 139K \\
26K+    & 21.4{\scriptsize$\pm$6.3} & 156K & 7.3K & 163K \\
\bottomrule
\end{tabular}
\caption{Inference statistics of \methodname{} (Qwen3-8B) per paper across length bins: agent steps and input, generated, and total tokens.}
\label{tab:length_cost}
\end{table}

Table~\ref{tab:length_bins} reports the distribution of papers across the five length bins.

\begin{table}[!ht]
\centering
\small
\begin{tabular}{@{}lcc@{}}
\toprule
\textbf{Bin} & \textbf{\# Papers} & \textbf{\% of Test Set} \\
\midrule
0--14K tokens & 169 & 16.9\% \\
14--18K tokens & 234 & 23.4\% \\
18--22K tokens & 202 & 20.2\% \\
22--26K tokens & 119 & 11.9\% \\
26K+ tokens & 276 & 27.6\% \\
\midrule
Total & 1{,}000 & 100\% \\
\bottomrule
\end{tabular}
\caption{Distribution of test papers across length bins.}
\label{tab:length_bins}
\end{table}

\subsection{Analysis of Undetected Injected Errors}\label{app:counterfactual_failures}

To understand why injected errors go undetected (\S\ref{sec:counterfactual}), we randomly sample 50 undetected cases per perturbation type (150 in total) and inspect each review log and final review.
The dominant cause differs by perturbation type.

\paragraph{Result perturbations: missed numerical changes.}
In about half of the cases, the agent does not re-check an altered value against the corresponding table or figure, especially when a single number is changed in a dense results table.
For example, when one entry in a results table is altered so that the method no longer outperforms a baseline, the agent records the text's claim of improvement as \texttt{supported} without comparing it with the modified entry.

\paragraph{Conclusion perturbations: identified but not reported.}
In 44\% of the cases, the review log records the inconsistency, but the final review prioritizes other weaknesses.
For example, the log contains a claim marked \texttt{weak} because the conclusion overstates the reported results, yet the final weaknesses list only missing baselines and limited ablations.
Better selection of log entries when composing the review could therefore recover part of these errors without further investigation.

\paragraph{Finding perturbations: causal-direction inversions.}
The hardest cases reverse the direction of a causal claim (12 of 50 cases), of which only 3 are mentioned, and only vaguely.
For example, a finding that ``A causes B'' is rewritten as ``B causes A''; since both statements are locally plausible, detecting the inversion requires reasoning about the causal structure implied by the experimental design rather than checking surface-level consistency.

\begin{table*}[!t]
  \centering
  \definecolor{bestcell}{HTML}{D5F5E3}
  \definecolor{secondcell}{HTML}{EBF5FB}
  \providecommand{\best}[1]{\cellcolor{bestcell}\textbf{#1}}
  \providecommand{\second}[1]{\cellcolor{secondcell}#1}
  \resizebox{\textwidth}{!}{
  \begin{tabular}{@{}llcccccccccc@{}}
  \toprule
  \multirow{2}{*}{\textbf{Method}} & \multirow{2}{*}{\textbf{Model}} &
  \multicolumn{2}{c}{\textbf{Actionability}} &
  \multicolumn{2}{c}{\textbf{Grounding}} &
  \multicolumn{2}{c}{\textbf{Technical Depth}} &
  \multicolumn{2}{c}{\textbf{Verifiability}} &
  \multicolumn{2}{c}{\textbf{Avg.}} \\
  \cmidrule(lr){3-4} \cmidrule(lr){5-6} \cmidrule(lr){7-8} \cmidrule(lr){9-10} \cmidrule(lr){11-12}
  & & avg@4 & best@4 & avg@4 & best@4 & avg@4 & best@4 & avg@4 & best@4 & avg@4 & best@4 \\
  \midrule

  \multicolumn{12}{c}{\textit{Prompt-based methods}} \\
  \midrule

  AgentReview & Llama3.1-8B & 0.18±0.10 & 0.19±0.10 & 0.07±0.09 & 0.09±0.10 & 0.08±0.08 & 0.09±0.09 & 0.16±0.11 & 0.18±0.11 & 0.12±0.10 & 0.14±0.10 \\
   & Qwen3-8B & 0.25±0.06 & 0.28±0.05 & 0.15±0.09 & 0.20±0.09 & 0.14±0.09 & 0.19±0.09 & 0.17±0.07 & 0.20±0.07 & 0.18±0.08 & 0.22±0.08 \\
   & Qwen3-14B & 0.26±0.06 & 0.30±0.06 & 0.14±0.09 & 0.20±0.10 & 0.17±0.10 & 0.24±0.09 & 0.24±0.07 & 0.26±0.06 & 0.20±0.08 & 0.25±0.08 \\
   & Qwen3-32B & 0.32±0.09 & 0.37±0.09 & 0.28±0.13 & 0.36±0.16 & 0.29±0.11 & 0.36±0.10 & 0.39±0.16 & 0.49±0.16 & 0.32±0.12 & 0.40±0.13 \\
   & Gemini-3.1-flash-lite & 0.36±0.09 & 0.41±0.09 & 0.31±0.12 & 0.39±0.12 & 0.34±0.11 & 0.40±0.10 & 0.37±0.12 & 0.45±0.11 & 0.34±0.11 & 0.42±0.10 \\
   & Qwen3.5-397B-A17B & 0.38±0.10 & 0.45±0.10 & \best{0.61±0.15} & \best{0.71±0.12} & 0.41±0.12 & 0.48±0.11 & 0.29±0.08 & 0.32±0.09 & 0.42±0.11 & 0.49±0.10 \\

  \cmidrule(lr){1-12}

  AI-Scientist-v2 & Llama3.1-8B & 0.21±0.09 & 0.26±0.08 & 0.09±0.11 & 0.15±0.12 & 0.08±0.09 & 0.12±0.10 & 0.18±0.09 & 0.21±0.08 & 0.14±0.10 & 0.18±0.09 \\
   & Qwen3-8B & 0.27±0.06 & 0.30±0.06 & 0.15±0.09 & 0.22±0.10 & 0.16±0.09 & 0.22±0.09 & 0.22±0.06 & 0.24±0.06 & 0.20±0.07 & 0.25±0.07 \\
   & Qwen3-14B & 0.26±0.07 & 0.31±0.06 & 0.13±0.10 & 0.19±0.10 & 0.16±0.10 & 0.22±0.09 & 0.24±0.06 & 0.27±0.06 & 0.20±0.08 & 0.25±0.08 \\
   & Qwen3-32B & 0.32±0.09 & 0.38±0.09 & 0.26±0.12 & 0.34±0.15 & 0.26±0.11 & 0.34±0.11 & 0.34±0.16 & 0.45±0.16 & 0.29±0.12 & 0.38±0.13 \\
   & Gemini-3.1-flash-lite & 0.33±0.09 & 0.39±0.09 & 0.26±0.12 & 0.34±0.12 & 0.32±0.13 & 0.40±0.12 & 0.27±0.11 & 0.34±0.12 & 0.29±0.11 & 0.37±0.11 \\
   & Qwen3.5-397B-A17B & 0.37±0.11 & 0.45±0.11 & 0.49±0.20 & 0.62±0.18 & 0.43±0.13 & 0.51±0.12 & 0.28±0.07 & 0.31±0.07 & 0.40±0.13 & 0.47±0.12 \\

  \cmidrule(lr){1-12}

  TreeReview & Llama3.1-8B & 0.25±0.04 & 0.27±0.04 & 0.33±0.12 & 0.42±0.11 & 0.24±0.10 & 0.30±0.09 & 0.14±0.11 & 0.21±0.12 & 0.24±0.09 & 0.30±0.09 \\
   & Qwen3-8B & 0.11±0.09 & 0.21±0.07 & 0.34±0.15 & 0.51±0.15 & 0.24±0.14 & 0.37±0.09 & 0.17±0.11 & 0.27±0.10 & 0.21±0.12 & 0.34±0.10 \\
   & Gemini-3.1-flash-lite & 0.34±0.10 & 0.43±0.11 & 0.52±0.12 & 0.57±0.10 & 0.36±0.12 & 0.44±0.13 & \best{0.52±0.12} & \best{0.56±0.11} & \second{0.44±0.12} & 0.50±0.11 \\
   & Qwen3.5-397B-A17B & 0.38±0.12 & 0.47±0.11 & \second{0.58±0.11} & \second{0.66±0.16} & 0.39±0.13 & 0.48±0.12 & 0.36±0.12 & 0.40±0.13 & 0.43±0.12 & 0.50±0.13 \\

  \midrule
  \multicolumn{12}{c}{\textit{Fine-tuned models}} \\
  \midrule

  CycleReviewer & Llama3.1-8B & 0.39±0.14 & 0.46±0.13 & 0.15±0.14 & 0.21±0.13 & 0.23±0.15 & 0.31±0.15 & 0.23±0.15 & 0.31±0.15 & 0.25±0.15 & 0.32±0.14 \\
   & Qwen3-8B & 0.27±0.12 & 0.32±0.14 & 0.20±0.18 & 0.30±0.21 & 0.18±0.14 & 0.24±0.14 & 0.25±0.12 & 0.28±0.13 & 0.22±0.14 & 0.28±0.16 \\
  DeepReview & Llama3.1-8B & 0.32±0.14 & 0.41±0.16 & 0.30±0.22 & 0.47±0.22 & 0.23±0.15 & 0.32±0.15 & 0.28±0.13 & 0.30±0.14 & 0.28±0.16 & 0.38±0.17 \\
   & Qwen3-8B & 0.36±0.14 & 0.44±0.12 & 0.12±0.14 & 0.21±0.15 & 0.20±0.14 & 0.29±0.14 & 0.20±0.16 & 0.29±0.16 & 0.22±0.15 & 0.31±0.14 \\
  ReviewRL (Reimpl.) & Qwen3-8B & \second{0.40±0.13} & \second{0.54±0.11} & 0.42±0.11 & 0.47±0.11 & 0.41±0.13 & 0.47±0.10 & 0.38±0.10 & 0.42±0.10 & 0.40±0.12 & 0.48±0.11 \\
  \addlinespace[2pt]
  \textbf{\methodname{} (ours)} & Llama3.1-8B & 0.38±0.13 & 0.48±0.12 & 0.50±0.21 & \second{0.66±0.19} & \best{0.46±0.14} & \best{0.55±0.12} & \second{0.43±0.14} & \second{0.51±0.13} & \second{0.44±0.15} & \second{0.55±0.14} \\
   & Qwen3-8B & \best{0.51±0.10} & \best{0.58±0.10} & 0.50±0.15 & 0.62±0.14 & \best{0.46±0.12} & \second{0.54±0.11} & 0.42±0.11 & 0.49±0.11 & \best{0.47±0.12} & \best{0.56±0.11} \\

  \bottomrule
  \end{tabular}
  }
  \caption{Evaluation results using GPT-5.4 nano as rubric judge (normalized to [0,1]). All scores are mean±std. \colorbox{bestcell}{\textbf{Green}} highlights the best and \colorbox{secondcell}{Blue} the second-best result in each column.}
  \label{tab:gpt54_results}
\end{table*}

\begin{table*}[!t]
  \centering
  \definecolor{bestcell}{HTML}{D5F5E3}
  \definecolor{secondcell}{HTML}{EBF5FB}
  \providecommand{\best}[1]{\cellcolor{bestcell}\textbf{#1}}
  \providecommand{\second}[1]{\cellcolor{secondcell}#1}
  \resizebox{\textwidth}{!}{
  \begin{tabular}{@{}llcccccccccc@{}}
  \toprule
  \multirow{2}{*}{\textbf{Method}} & \multirow{2}{*}{\textbf{Model}} &
  \multicolumn{2}{c}{\textbf{Actionability}} &
  \multicolumn{2}{c}{\textbf{Grounding}} &
  \multicolumn{2}{c}{\textbf{Technical Depth}} &
  \multicolumn{2}{c}{\textbf{Verifiability}} &
  \multicolumn{2}{c}{\textbf{Avg.}} \\
  \cmidrule(lr){3-4} \cmidrule(lr){5-6} \cmidrule(lr){7-8} \cmidrule(lr){9-10} \cmidrule(lr){11-12}
  & & avg@4 & best@4 & avg@4 & best@4 & avg@4 & best@4 & avg@4 & best@4 & avg@4 & best@4 \\
  \midrule

  \multicolumn{12}{c}{\textit{Prompt-based methods}} \\
  \midrule

  AgentReview & Llama3.1-8B & 0.19±0.09 & 0.20±0.09 & 0.26±0.12 & 0.28±0.13 & 0.18±0.11 & 0.20±0.11 & 0.02±0.05 & 0.03±0.06 & 0.16±0.09 & 0.18±0.10 \\
   & Qwen3-8B & 0.26±0.05 & 0.29±0.05 & 0.36±0.09 & 0.42±0.09 & 0.27±0.09 & 0.34±0.09 & 0.02±0.05 & 0.04±0.08 & 0.23±0.07 & 0.27±0.08 \\
   & Qwen3-14B & 0.28±0.05 & 0.30±0.06 & 0.40±0.09 & 0.45±0.09 & 0.31±0.10 & 0.37±0.10 & 0.09±0.11 & 0.17±0.13 & 0.27±0.09 & 0.32±0.10 \\
   & Qwen3-32B & 0.28±0.05 & 0.30±0.06 & 0.42±0.09 & 0.47±0.09 & 0.33±0.10 & 0.39±0.10 & 0.09±0.10 & 0.16±0.12 & 0.28±0.09 & 0.33±0.09 \\
   & Gemini-3.1-flash-lite & 0.36±0.09 & 0.42±0.11 & 0.53±0.10 & 0.60±0.11 & \best{0.53±0.13} & \best{0.60±0.12} & 0.34±0.18 & 0.48±0.16 & 0.44±0.12 & 0.52±0.12 \\
   & Qwen3.5-397B-A17B & 0.35±0.10 & 0.41±0.12 & 0.54±0.11 & 0.61±0.12 & 0.49±0.14 & 0.57±0.13 & 0.38±0.17 & 0.50±0.14 & 0.44±0.13 & 0.52±0.13 \\

  \cmidrule(lr){1-12}

  AI-Scientist-v2 & Llama3.1-8B & 0.21±0.08 & 0.26±0.06 & 0.29±0.10 & 0.37±0.10 & 0.19±0.10 & 0.27±0.10 & 0.02±0.06 & 0.05±0.09 & 0.18±0.08 & 0.24±0.09 \\
   & Qwen3-8B & 0.27±0.05 & 0.29±0.06 & 0.36±0.09 & 0.42±0.09 & 0.27±0.09 & 0.33±0.09 & 0.06±0.09 & 0.12±0.11 & 0.24±0.08 & 0.29±0.09 \\
   & Qwen3-14B & 0.27±0.06 & 0.30±0.06 & 0.37±0.10 & 0.43±0.10 & 0.28±0.10 & 0.34±0.10 & 0.10±0.12 & 0.18±0.13 & 0.26±0.09 & 0.32±0.10 \\
   & Qwen3-32B & 0.27±0.05 & 0.30±0.06 & 0.40±0.09 & 0.45±0.09 & 0.30±0.10 & 0.37±0.10 & 0.08±0.10 & 0.15±0.12 & 0.26±0.08 & 0.32±0.09 \\
   & Gemini-3.1-flash-lite & 0.31±0.09 & 0.36±0.10 & 0.47±0.10 & 0.52±0.10 & 0.45±0.15 & 0.54±0.14 & 0.28±0.16 & 0.39±0.14 & 0.38±0.12 & 0.45±0.12 \\
   & Qwen3.5-397B-A17B & 0.36±0.10 & 0.42±0.11 & 0.54±0.11 & 0.61±0.11 & 0.46±0.14 & 0.52±0.13 & 0.43±0.14 & 0.53±0.12 & 0.45±0.12 & 0.51±0.12 \\

  \cmidrule(lr){1-12}

  TreeReview & Llama3.1-8B & 0.19±0.11 & 0.29±0.09 & 0.32±0.19 & 0.42±0.17 & 0.14±0.13 & 0.20±0.16 & 0.02±0.06 & 0.08±0.06 & 0.17±0.12 & 0.25±0.12 \\
   & Qwen3-8B & 0.09±0.10 & 0.20±0.10 & 0.45±0.12 & 0.58±0.06 & 0.32±0.11 & 0.44±0.07 & 0.16±0.15 & 0.31±0.11 & 0.26±0.12 & 0.38±0.08 \\
   & Gemini-3.1-flash-lite & 0.33±0.12 & 0.41±0.11 & \best{0.67±0.10} & \second{0.69±0.12} & 0.44±0.11 & 0.52±0.12 & 0.45±0.12 & 0.51±0.09 & 0.47±0.11 & 0.53±0.11 \\
   & Qwen3.5-397B-A17B & \best{0.40±0.10} & 0.42±0.12 & \second{0.62±0.11} & 0.65±0.12 & 0.45±0.15 & 0.52±0.15 & \best{0.62±0.16} & \second{0.63±0.16} & \second{0.52±0.13} & \second{0.56±0.14} \\

  \midrule
  \multicolumn{12}{c}{\textit{Fine-tuned models}} \\
  \midrule

  CycleReviewer & Llama3.1-8B & 0.35±0.17 & \second{0.44±0.13} & 0.28±0.13 & 0.34±0.15 & 0.28±0.15 & 0.35±0.18 & 0.08±0.18 & 0.17±0.25 & 0.25±0.16 & 0.32±0.18 \\
   & Qwen3-8B & 0.29±0.14 & 0.35±0.14 & 0.33±0.14 & 0.39±0.13 & 0.25±0.14 & 0.30±0.16 & 0.14±0.21 & 0.17±0.23 & 0.25±0.16 & 0.30±0.17 \\
  DeepReview & Llama3.1-8B & 0.32±0.14 & 0.40±0.15 & 0.39±0.15 & 0.49±0.14 & 0.28±0.16 & 0.39±0.20 & 0.28±0.25 & 0.33±0.27 & 0.32±0.17 & 0.40±0.19 \\
   & Qwen3-8B & 0.32±0.16 & 0.42±0.13 & 0.27±0.13 & 0.34±0.16 & 0.26±0.15 & 0.36±0.18 & 0.06±0.15 & 0.15±0.23 & 0.23±0.15 & 0.32±0.17 \\
  ReviewRL (Reimpl.) & Qwen3-8B & 0.30±0.15 & \second{0.44±0.12} & 0.45±0.15 & 0.58±0.12 & 0.31±0.19 & 0.47±0.12 & 0.53±0.12 & 0.57±0.11 & 0.40±0.15 & 0.52±0.12 \\
  \addlinespace[2pt]
  \textbf{\methodname{} (ours)} & Llama3.1-8B & 0.33±0.08 & 0.35±0.09 & 0.53±0.15 & 0.63±0.14 & 0.42±0.15 & 0.53±0.14 & 0.51±0.17 & 0.61±0.14 & 0.45±0.14 & 0.53±0.13 \\
   & Qwen3-8B & \second{0.39±0.12} & \best{0.47±0.14} & \second{0.62±0.14} & \best{0.73±0.12} & \second{0.49±0.14} & \second{0.59±0.14} & \second{0.61±0.13} & \best{0.69±0.11} & \best{0.53±0.13} & \best{0.62±0.13} \\

  \bottomrule
  \end{tabular}
  }
  \caption{Evaluation results using DeepSeek-V4 flash as rubric judge (normalized to [0,1]). All scores are mean±std. \colorbox{bestcell}{\textbf{Green}} highlights the best and \colorbox{secondcell}{Blue} the second-best result in each column.}
  \label{tab:deepseekv4_results}
\end{table*}

\begin{table*}[!t]
  \centering
  \definecolor{bestcell}{HTML}{D5F5E3}
  \definecolor{secondcell}{HTML}{EBF5FB}
  \providecommand{\best}[1]{\cellcolor{bestcell}\textbf{#1}}
  \providecommand{\second}[1]{\cellcolor{secondcell}#1}
  \resizebox{\textwidth}{!}{
  \begin{tabular}{@{}llcccccccccc@{}}
  \toprule
  \multirow{2}{*}{\textbf{Method}} & \multirow{2}{*}{\textbf{Model}} &
  \multicolumn{2}{c}{\textbf{Actionability}} &
  \multicolumn{2}{c}{\textbf{Grounding}} &
  \multicolumn{2}{c}{\textbf{Technical Depth}} &
  \multicolumn{2}{c}{\textbf{Verifiability}} &
  \multicolumn{2}{c}{\textbf{Avg.}} \\
  \cmidrule(lr){3-4} \cmidrule(lr){5-6} \cmidrule(lr){7-8} \cmidrule(lr){9-10} \cmidrule(lr){11-12}
  & & avg@4 & best@4 & avg@4 & best@4 & avg@4 & best@4 & avg@4 & best@4 & avg@4 & best@4 \\
  \midrule

  \multicolumn{12}{c}{\textit{Prompt-based methods}} \\
  \midrule

  AgentReview & Llama3.1-8B & 0.02±0.06 & 0.04±0.07 & 0.02±0.05 & 0.03±0.07 & --- & --- & 0.00±0.00 & 0.00±0.00 & 0.01±0.04 & 0.02±0.05 \\
   & Qwen3-8B & 0.38±0.14 & 0.49±0.13 & 0.45±0.18 & 0.59±0.14 & --- & --- & 0.22±0.14 & 0.31±0.16 & 0.35±0.15 & 0.47±0.15 \\
   & Qwen3-14B & 0.04±0.08 & 0.08±0.10 & 0.07±0.11 & 0.13±0.13 & --- & --- & 0.00±0.01 & 0.00±0.02 & 0.04±0.06 & 0.07±0.08 \\
   & Qwen3-32B & 0.02±0.05 & 0.05±0.08 & 0.05±0.08 & 0.12±0.11 & --- & --- & 0.00±0.01 & 0.00±0.03 & 0.02±0.05 & 0.06±0.07 \\
   & Gemini-3.1-flash-lite & 0.16±0.15 & 0.29±0.16 & 0.25±0.19 & 0.40±0.17 & --- & --- & 0.03±0.07 & 0.06±0.10 & 0.15±0.14 & 0.25±0.14 \\
   & Qwen3.5-397B-A17B & 0.10±0.12 & 0.20±0.14 & 0.23±0.19 & 0.38±0.19 & --- & --- & 0.02±0.06 & 0.05±0.08 & 0.12±0.12 & 0.21±0.14 \\

  \cmidrule(lr){1-12}

  AI-Scientist-v2 & Llama3.1-8B & 0.08±0.15 & 0.22±0.19 & 0.10±0.19 & 0.28±0.25 & --- & --- & 0.00±0.01 & 0.00±0.02 & 0.06±0.12 & 0.17±0.16 \\
   & Qwen3-8B & 0.08±0.10 & 0.17±0.12 & 0.10±0.12 & 0.21±0.14 & --- & --- & 0.00±0.02 & 0.01±0.03 & 0.06±0.08 & 0.13±0.10 \\
   & Qwen3-14B & 0.09±0.11 & 0.19±0.13 & 0.12±0.14 & 0.24±0.15 & --- & --- & 0.00±0.02 & 0.01±0.04 & 0.07±0.09 & 0.14±0.11 \\
   & Qwen3-32B & 0.04±0.08 & 0.10±0.10 & 0.08±0.11 & 0.18±0.13 & --- & --- & 0.00±0.02 & 0.01±0.04 & 0.04±0.07 & 0.10±0.09 \\
   & Gemini-3.1-flash-lite & 0.14±0.16 & 0.27±0.17 & 0.19±0.18 & 0.35±0.18 & --- & --- & 0.02±0.07 & 0.05±0.10 & 0.12±0.13 & 0.22±0.15 \\
   & Qwen3.5-397B-A17B & 0.12±0.12 & 0.21±0.13 & 0.29±0.18 & 0.43±0.17 & --- & --- & 0.04±0.07 & 0.08±0.09 & 0.15±0.12 & 0.24±0.13 \\

  \cmidrule(lr){1-12}

  TreeReview & Llama3.1-8B & 0.22±0.22 & 0.42±0.24 & 0.17±0.17 & 0.33±0.14 & --- & --- & 0.04±0.15 & 0.19±0.18 & 0.14±0.18 & 0.31±0.19 \\
   & Qwen3-8B & 0.30±0.14 & 0.43±0.13 & 0.39±0.17 & 0.54±0.15 & --- & --- & 0.19±0.09 & 0.25±0.01 & 0.29±0.13 & 0.41±0.10 \\
   & Gemini-3.1-flash-lite & 0.20±0.15 & 0.33±0.18 & 0.46±0.16 & 0.54±0.17 & --- & --- & \second{0.34±0.07} & 0.36±0.09 & 0.33±0.13 & 0.41±0.15 \\
   & Qwen3.5-397B-A17B & 0.27±0.15 & 0.31±0.13 & \second{0.78±0.17} & \second{0.85±0.15} & --- & --- & \best{0.55±0.14} & \best{0.56±0.15} & \best{0.53±0.15} & \second{0.57±0.14} \\

  \midrule
  \multicolumn{12}{c}{\textit{Fine-tuned models}} \\
  \midrule

  CycleReviewer & Llama3.1-8B & \second{0.41±0.32} & 0.57±0.30 & 0.32±0.27 & 0.45±0.24 & --- & --- & 0.12±0.22 & 0.23±0.25 & 0.29±0.27 & 0.42±0.26 \\
   & Qwen3-8B & 0.37±0.30 & 0.45±0.30 & 0.31±0.26 & 0.39±0.26 & --- & --- & 0.13±0.22 & 0.19±0.24 & 0.27±0.26 & 0.34±0.27 \\
  DeepReview & Llama3.1-8B & 0.39±0.29 & \best{0.60±0.25} & 0.35±0.27 & 0.55±0.23 & --- & --- & 0.17±0.23 & 0.35±0.23 & 0.30±0.26 & 0.50±0.24 \\
   & Qwen3-8B & 0.36±0.29 & 0.53±0.26 & 0.26±0.26 & 0.43±0.24 & --- & --- & 0.12±0.21 & 0.26±0.25 & 0.25±0.25 & 0.41±0.25 \\
  ReviewRL (Reimpl.) & Qwen3-8B & 0.28±0.21 & 0.49±0.20 & 0.67±0.20 & 0.77±0.16 & --- & --- & 0.22±0.16 & 0.30±0.15 & 0.39±0.19 & 0.52±0.17 \\
  \addlinespace[2pt]
  \textbf{\methodname{} (ours)} & Llama3.1-8B & 0.20±0.19 & 0.36±0.18 & 0.66±0.21 & 0.81±0.16 & --- & --- & 0.27±0.18 & \second{0.38±0.15} & 0.38±0.19 & 0.52±0.16 \\
   & Qwen3-8B & \best{0.47±0.16} & \second{0.59±0.14} & \best{0.80±0.18} & \best{0.92±0.10} & --- & --- & 0.16±0.15 & 0.27±0.15 & \second{0.48±0.16} & \best{0.59±0.13} \\

  \bottomrule
  \end{tabular}
  }
  \caption{Evaluation results using the RevUtil judge. The trained RevUtil can only score Actionability, Grounding, and Verifiability, so Technical Depth is marked ``---''. All scores are mean±std (on [0,1] scale). \colorbox{bestcell}{\textbf{Green}} highlights the best and \colorbox{secondcell}{Blue} the second-best result in each column.}
  \label{tab:utility_results}
\end{table*}

\begin{table*}[!t]
  \centering
  \providecommand{\ms}[2]{#1{\scriptsize\textcolor{black!90}{$\pm$#2}}}
  \resizebox{\textwidth}{!}{
  \begin{tabular}{@{}llcccccccccccc@{}}
  \toprule
  \multirow{2}{*}{\textbf{Method}} & \multirow{2}{*}{\textbf{Model}} &
  \multicolumn{2}{c}{\textbf{Actionability}} &
  \multicolumn{2}{c}{\textbf{Grounding}} &
  \multicolumn{2}{c}{\textbf{Technical Depth}} &
  \multicolumn{2}{c}{\textbf{Verifiability}} &
  \multicolumn{2}{c}{\textbf{Score Alignment}} &
  \multicolumn{2}{c}{\textbf{Avg.}} \\
  \cmidrule(lr){3-4} \cmidrule(lr){5-6} \cmidrule(lr){7-8} \cmidrule(lr){9-10} \cmidrule(lr){11-12} \cmidrule(lr){13-14}
  & & avg@4 & best@4 & avg@4 & best@4 & avg@4 & best@4 & avg@4 & best@4 & avg@4 & best@4 & avg@4 & best@4 \\
  \midrule
  AgentReview& Qwen3-14B & \ms{0.19}{0.06} & \ms{0.22}{0.07} & \ms{0.20}{0.10} & \ms{0.26}{0.10} & \ms{0.24}{0.10} & \ms{0.30}{0.10} & \ms{0.11}{0.06} & \ms{0.15}{0.07} & \ms{0.61}{0.13} & \ms{0.63}{0.13} & \ms{0.27}{0.09} & \ms{0.31}{0.09} \\
   & Qwen3-32B & \ms{0.20}{0.06} & \ms{0.24}{0.07} & \ms{0.25}{0.10} & \ms{0.32}{0.12} & \ms{0.31}{0.11} & \ms{0.38}{0.10} & \ms{0.16}{0.09} & \ms{0.22}{0.10} & \ms{0.64}{0.13} & \ms{0.69}{0.13} & \ms{0.31}{0.10} & \ms{0.37}{0.11} \\
  \cmidrule(lr){1-14}
  AI-Scientist-v2& Qwen3-14B & \ms{0.21}{0.08} & \ms{0.27}{0.09} & \ms{0.21}{0.11} & \ms{0.29}{0.12} & \ms{0.22}{0.10} & \ms{0.28}{0.10} & \ms{0.11}{0.07} & \ms{0.15}{0.08} & \ms{0.69}{0.13} & \ms{0.73}{0.13} & \ms{0.29}{0.10} & \ms{0.34}{0.10} \\
   & Qwen3-32B & \ms{0.21}{0.07} & \ms{0.26}{0.08} & \ms{0.24}{0.10} & \ms{0.33}{0.12} & \ms{0.28}{0.10} & \ms{0.35}{0.10} & \ms{0.14}{0.09} & \ms{0.20}{0.10} & \ms{0.69}{0.13} & \ms{0.73}{0.12} & \ms{0.31}{0.10} & \ms{0.37}{0.11} \\
  \bottomrule
  \end{tabular}
  }
  \caption{Results of prompt-based baselines with intermediate-size backbones, averaged across three judges (DeepSeek-V4 flash, GPT-5.4 nano, RevUtil). All scores are mean{\scriptsize$\pm$std} (on [0,1] scale) over four independent reviews per paper; Technical Depth is averaged over the two judges that support it.}
  \label{tab:mid_size}
\end{table*}

\section{Inference Cost Analysis}\label{app:complexity}

To better understand the inference cost of \methodname{}, which makes multiple LLM calls per paper as a multi-step agent, we provide an empirical comparison with the baselines in terms of token usage, inference time, and monetary cost, as well as review quality under an equal token budget.
Unless otherwise noted, \methodname{} refers to the trained Qwen3-8B model.

\paragraph{\methodname{} uses more tokens than methods that place the full paper in context, such as DeepReview, but far fewer than TreeReview.}
Single-pass and multi-stage methods read the full paper (12--20K tokens) in each of their 1--5 calls, using ${\sim}$20K and ${\sim}$60K tokens per paper, respectively.
As \methodname{} is a sequential agentic method, it uses ${\sim}$135K tokens per paper on average over about 20 steps.
TreeReview uses far more tokens, as it issues LLM calls for each node of its question tree (85--102 nodes on average; Appendix~\ref{app:paper_length}): ${\sim}$630K per paper with the same Qwen3-8B backbone, about 4.7 times as many as \methodname{}, and ${\sim}$550K with Qwen3.5-397B-A17B.

\paragraph{\methodname{} is faster than TreeReview and comparable to DeepReview.}
Run sequentially, a single \methodname{} review takes ${\sim}$1--2 minutes on a single A100 (80\,GiB) GPU, compared with 5--6 minutes for TreeReview with the same Qwen3-8B backbone and ${\sim}$16 minutes with Qwen3.5-397B-A17B; latency remains moderate despite the number of calls because each step generates only a short structured action.
When the model is served with vLLM on an 8$\times$A100 (80\,GiB) node with papers processed concurrently, \methodname{} reaches a throughput of ${\sim}$32 seconds per paper, comparable to DeepReview (6--45 seconds).

\paragraph{Under public API prices, a \methodname{} review costs less than one cent, far below TreeReview.}
Using Groq API pricing for Llama3.1-8B as a representative price for an 8B model (\$0.05 and \$0.08 per million input and output tokens), a \methodname{} review (${\sim}$129K input and ${\sim}$6.8K output tokens on average) costs ${\sim}$\$0.007, compared with ${\sim}$\$0.001 for AgentReview, ${\sim}$\$0.003 for DeepReview, and ${\sim}$\$0.03 for TreeReview with the same backbone, about 4.5 times as much as \methodname{}.
With Qwen3.5-397B-A17B, TreeReview costs ${\sim}$\$0.50 per paper at public API pricing for this model.

\paragraph{Even when cheaper baselines are given multiple attempts, \methodname{} remains ahead.}
We give each baseline approximately the same token budget (${\sim}$135K tokens per paper): cheaper methods are run multiple times and we report their best result, while \methodname{} is run once.
As shown in Table~\ref{tab:equal_budget}, \methodname{} achieves the highest score (0.57), so its gains are not explained by its larger inference budget.
TreeReview already exceeds this budget with a single run (480--625K tokens) yet scores lower (0.55).

\begin{table}[!ht]
\centering
\small
\resizebox{\columnwidth}{!}{%
\begin{tabular}{@{}lccc@{}}
\toprule
\textbf{Method} & \textbf{Tokens/run} & \textbf{Runs} & \textbf{Avg.\ Score} \\
\midrule
AgentReview (397B)     & ${\sim}$20K & Best of 6 & 0.50 \\
AI-Scientist-v2 (397B) & ${\sim}$60K & Best of 2 & 0.48 \\
TreeReview (397B)      & ${\sim}$480--625K & Single run & 0.55 \\
DeepReview (8B)        & ${\sim}$60K & Best of 2 & 0.40 \\
\methodname{} (8B)      & ${\sim}$135K & Single run & \textbf{0.57} \\
\bottomrule
\end{tabular}}
\caption{Comparison under a matched inference budget of ${\sim}$135K tokens per paper. TreeReview exceeds this budget with a single run; we report its single-run score for reference. 397B = Qwen3.5-397B-A17B; 8B = Qwen3-8B.}
\label{tab:equal_budget}
\end{table}

\section{Case Study}\label{app:case_study}

To illustrate how the evolving review log guides investigation and links the final review to the collected evidence, we present \methodname{}'s review of an ICLR submission titled \textit{``Surf3R: Rapid Surface Reconstruction from Sparse RGB Views in Seconds.''}
The agent completed the review in 30 steps, accumulating 12 claims, 6 questions, and 32 notes in its review log.
Figure~\ref{fig:case_trajectory} shows the abridged trajectory: the agent flags suspicious claims early, cross-references them against experimental evidence, updates their status as new information emerges, and ultimately derives each review point from specific log entries---demonstrating the traceability and proactive investigation.

\begin{figure*}[t!]
\centering
\small

\begin{tcolorbox}[casestyle, title={\textbf{Step 1}}]
\footnotesize
\textbf{Env action:} \texttt{read\_section(abstract)}\\
\textbf{Observation:} \textit{``\ldots completes an entire scene in under 10 seconds \ldots multi-branch and multi-view decoding architecture \ldots D-Normal regularizer \ldots''}\\
\textbf{Log action:} \texttt{log(claim, C1:} \textit{``Surf3R completes an entire scene in under 10 seconds.''} \textsf{to\_be\_verified}\texttt{)}\\
\hphantom{\textbf{Log action:}} \texttt{log(note, N2:} \textit{``10-second claim doesn't specify how many views or what hardware. Need to verify.''}\texttt{)}\\
\hphantom{\textbf{Log action:}} \texttt{log(question, Q1:} \textit{``Does the multi-branch architecture improve quality vs.\ a single-reference baseline?''} \textsf{open}\texttt{)}\\
\hphantom{\textbf{Log action:}} \texttt{log(question, Q2:} \textit{``Does the paper ablate D-Normal vs.\ other regularization methods?''} \textsf{open}\texttt{)}
\end{tcolorbox}
\vspace{-4pt}

\begin{tcolorbox}[omitstyle]
\footnotesize
\textbf{\ldots\; Step 2--5:}\; \texttt{read\_section(introduction)}, \texttt{read\_section(methodology)}, \texttt{read\_section(experiments)}, \texttt{read\_section(ablation\_study)} --- \textit{logs claims C2--C12 about architecture, loss design, and ablation results.}
\end{tcolorbox}
\vspace{-4pt}

\begin{tcolorbox}[casestyle, title={\textbf{Step 6}}]
\footnotesize
\textbf{Env action:} \texttt{read\_section(ablation\_study)} \textit{(re-read)}\\
\textbf{Observation:} \textit{Table~5 Row~A: removing multi-branch drops F1 36.66 $\to$ 23.24.\; Row~D: removing D-Normal drops F1 41.92 $\to$ 30.96.}\\
\textbf{Log action:} \texttt{log(note, N8:} \textit{``Multi-branch ablation is within the same model---no separately trained single-branch baseline.''}\texttt{)}\\
\hphantom{\textbf{Log action:}} \texttt{update(Q1,} \textsf{open} $\to$ \textsf{partially\_answered} --- \textit{ablation exists but within-model only}\texttt{)}\\
\hphantom{\textbf{Log action:}} \texttt{update(Q2,} \textsf{open} $\to$ \textsf{partially\_answered} --- \textit{component ablation but no alternative methods}\texttt{)}
\end{tcolorbox}
\vspace{-4pt}

\begin{tcolorbox}[omitstyle]
\footnotesize
\textbf{\ldots\; Step 7--22:}\; \texttt{look\_up(``10 seconds'')}, \texttt{look\_up(``GPU'')}, \texttt{look\_up(``inference time'')}, \texttt{read\_section(implementation\_details)}, \ldots --- \textit{finds 32 H800 GPUs for training but no inference hardware or timing reported.}
\end{tcolorbox}
\vspace{-4pt}

\begin{tcolorbox}[casestyle, title={\textbf{Step 23}}]
\footnotesize
\textbf{Env action:} \texttt{read\_section(NVS\_results)}\\
\textbf{Observation:} \textit{Table~3 compares Surf3R-GD only against DUSt3R (PSNR 15.06 vs.\ 11.66 at 4 views). No other NVS baselines.}\\
\textbf{Log action:} \texttt{log(note, N9:} \textit{``NVS evaluation compares only against DUSt3R. No other baselines provided.''}\texttt{)}\\
\hphantom{\textbf{Log action:}} \texttt{update(C1,} \textsf{to\_be\_verified} $\to$ \textsf{invalid} --- \textit{no inference hardware or timing found}\texttt{)}\\
\hphantom{\textbf{Log action:}} \texttt{update(C4,} \textsf{to\_be\_verified} $\to$ \textsf{weak} --- \textit{SOTA claimed but NVS lacks baselines beyond DUSt3R}\texttt{)}
\end{tcolorbox}
\vspace{-4pt}

\begin{tcolorbox}[omitstyle]
\footnotesize
\textbf{\ldots\; Step 24--29:}\; \texttt{read\_section(surface\_reconstruction)}, \texttt{read\_section(conclusion)}, \ldots --- \textit{finds 180$\times$ speedup claim without specifying baseline; logs N29.}
\end{tcolorbox}
\vspace{-4pt}

\begin{tcolorbox}[resultstyle, title={\textbf{Step 30: Review generation}}]
\footnotesize
\textbf{Env action:} \texttt{finish}\\
\textbf{Log action:} \texttt{outline(summary)} $\to$ \texttt{outline(strengths)} $\to$ \texttt{outline(weaknesses)} $\to$ \texttt{outline(questions)} $\to$ \texttt{outline(score)}\\[2pt]
\textbf{Summary:} \textit{``Surf3R proposes a feed-forward framework for pose-free 3D surface reconstruction from sparse RGB views using a multi-branch architecture and a D-Normal regularizer based on 3D Gaussians \ldots''}\\[3pt]
\textbf{Strength 1} \texttt{[C5, C6, N4]}: \textit{``Novel feed-forward approach for pose-free surface reconstruction \ldots''}\\
\textbf{Strength 2} \texttt{[C7, C10, C11]}: \textit{``D-Normal Regularizer confirmed by ablation (Table~5, Row~D: F1 41.92 $\to$ 30.96) \ldots''}\\[3pt]
\textbf{Weakness 1} \texttt{[C1, N2, N26, N28]}: \textit{``The paper claims completion in under 10 seconds (abstract, introduction) but provides no hardware specifications (number of GPUs, inference time) or number of views. Section~4.1 mentions 32 H800 GPUs for training but does not specify inference hardware.''}\\[2pt]
\textbf{Weakness 2} \texttt{[C2, Q1, N8]}: \textit{``Table~5 Row~A shows that removing multi-branch degrades F1 from 36.66 to 23.24, but this is within the same model. A proper ablation would train a single-branch model separately to confirm the architecture's necessity.''}\\[2pt]
\textbf{Weakness 3--5} \ldots\hfill \textit{(3 more weaknesses about D-Normal ablation, NVS baselines, and speedup claim)}\\[3pt]
\textbf{Overall score: 5/10}
\end{tcolorbox}

\caption{Abridged review trajectory of \methodname{} on \textit{Surf3R} (30 steps; 12 claims, 6 questions, 32 notes). Orange boxes show key steps; gray boxes summarize omitted steps with ``\ldots''; the green box shows the final review. Each review point traces back to evidence entries accumulated during investigation.}
\label{fig:case_trajectory}
\end{figure*}

\section{Prototype: External Search for Novelty Assessment}\label{app:novelty_search}

We demonstrate how \methodname{} can be extended to incorporate external evidence.
Specifically, we introduce a single environment action, \texttt{novelty\_checker(claim)}, which is executed by a Novelty Checker subagent.
Whenever \methodname{} encounters a novelty claim, it invokes this subagent to assess the claim against prior work.
The subagent can be implemented using any novelty-checking method.
In our prototype, the subagent
(1)~reformulates the claim into keyword queries,
(2)~retrieves the top-$k$ candidates (title, abstract, year) via the arXiv API\footnote{\texttt{export.arxiv.org/api/query}}, restricted to papers published before the submission date,
(3)~filters candidates for topical overlap with the claim, and
(4)~synthesizes a compact report: the 3--5 most related works, each with a one-line overlap note, plus an overall novelty verdict (\texttt{novel} $|$ \texttt{partial\_overlap} $|$ \texttt{known}).

The resulting report is returned as a standard observation.
The main agent records it as a note and updates the status of the corresponding novelty claim accordingly.
Importantly, the action interface specifies only the input (a claim) and output (a compact report).

\paragraph{Example.}
Figure~\ref{fig:search_prototype} illustrates the interaction.
The main agent records the paper's novelty claim, delegates it to the Novelty Checker, and, after receiving evidence of closely related prior work, downgrades the claim and produces an externally grounded weakness.

\begin{figure*}[t!]
\centering
\small

\begin{tcolorbox}[casestyle, title={\textbf{Delegation to the Novelty Checker}}]
\footnotesize
\textbf{Review log:} \texttt{C13:} \textit{``First feed-forward framework for pose-free surface reconstruction from sparse RGB views.''} \textsf{to\_be\_verified}\\[2pt]
\textbf{Env action:} \texttt{novelty\_checker(claim=C13)}\\
\textbf{Observation:}
\textit{(i) Splatt3R (2024): feed-forward, pose-free 3D Gaussian reconstruction from uncalibrated image pairs---same setting, but targets novel-view rendering rather than high-quality surface reconstruction.
(ii) SparseNeuS (2022): generalizable surface reconstruction from sparse views---requires camera poses.
Verdict: \textsf{partial\_overlap}.}\\[2pt]
\textbf{Log action:} \texttt{log(note, N33:} \textit{``Feed-forward pose-free reconstruction already exists (Splatt3R); the claimed novelty is limited to the surface reconstruction setting.''}\texttt{)}\\
\hphantom{\textbf{Log action:}} \texttt{update(C13,} \textsf{to\_be\_verified} $\rightarrow$ \textsf{weak} --- \textit{prior work already covers the pose-free feed-forward setting}\texttt{)}
\end{tcolorbox}
\vspace{-4pt}

\begin{tcolorbox}[resultstyle, title={\textbf{Review outcome}}]
\footnotesize
\textbf{Weakness} \texttt{[C13, N33]}: \textit{``The claim of being the first feed-forward pose-free framework is overstated: Splatt3R already performs feed-forward pose-free reconstruction from uncalibrated images. The true novelty is the surface reconstruction setting, which should be stated explicitly.''}
\end{tcolorbox}

\caption{Prototype interaction with the Novelty Checker on the \textit{Surf3R} example. The main agent delegates the novelty claim (\texttt{C13}) to the Novelty Checker, whose compact report on related prior work leads the agent to downgrade the claim and generate an externally grounded weakness.}
\label{fig:search_prototype}
\end{figure*}

\end{document}